\documentclass{article} 
\usepackage{iclr2026_conference,times}

\usepackage[utf8]{inputenc} 
\usepackage[T1]{fontenc}    
\usepackage{hyperref}       
\usepackage{url}            
\usepackage{booktabs}       
\usepackage{amsfonts}       
\usepackage{nicefrac}       
\usepackage{microtype}      
\usepackage[table]{xcolor}         

\usepackage{listings}
\usepackage{amsmath}
\usepackage{mathtools}
\usepackage[ruled, vlined, linesnumbered]{algorithm2e}
\SetAlFnt{\small}
\SetAlCapFnt{\small}
\SetAlCapNameFnt{\small}
\usepackage{multirow}
\usepackage{graphicx}
\usepackage{makecell}
\usepackage{caption}
\usepackage{subcaption}
\usepackage{wrapfig}
\usepackage{mathpartir}
\usepackage{adjustbox}
\usepackage{mdframed}
\usepackage{siunitx}
\usepackage{pifont}
\newcommand{\xmark}{\ding{55}}%

\newcommand{\rcmark}{\textcolor{red}{\textbf{\ding{51}}}}
\usepackage[cachedir=minted-cache]{minted}
\usepackage{dsfont}
\usepackage{amssymb}
\usepackage{tcolorbox}
\usepackage[hybrid,texMathDollars]{markdown}
\DeclareUnicodeCharacter{2081}{$_1$}
\DeclareUnicodeCharacter{2082}{$_2$}
\DeclareUnicodeCharacter{2099}{$_{\mathrm n}$}
\usepackage{etoolbox}
\usepackage{tocloft}
\usepackage{appendix}
\usepackage{titletoc}

\newcommand{\Name}{{HeuriGym}\xspace}

\newcommand{\RNum}[1]{\uppercase\expandafter{\romannumeral #1\relax}}
\newcommand{\passk}{\textsc{pass@$k$}\xspace}
\newcommand{\solvesi}{\textsc{solve$_s$@$i$}}
\newcommand{\solve}[2]{\textsc{solve$_\text{\RNum{#1}}$@${#2}$}}

\newcommand{\GPTmini}{\textsf{GPT-o4-mini-high}\xspace}
\newcommand{\Claude}{\textsf{Claude-3.7-Sonnet}\xspace}
\newcommand{\DeepSeekV}{\textsf{DeepSeek-V3}\xspace}
\newcommand{\DeepSeekR}{\textsf{DeepSeek-R1}\xspace}
\newcommand{\GeminiFlash}{\textsf{Gemini-2.5-Flash}\xspace}
\newcommand{\GeminiPro}{\textsf{Gemini-2.5-Pro}\xspace}
\newcommand{\Llamas}{\textsf{LLaMA-3.3}\xspace}
\newcommand{\Llamal}{\textsf{LLaMA-4-Maverick}\xspace}
\newcommand{\Qwen}{\textsf{Qwen3-235B}\xspace}

\newcommand{\showcomments}{yes}
\newcommand\fixme[1]{
    \ifthenelse{\equal{\showcomments}{yes}}{{\textcolor{red}{\small [FIXME: #1]}}}{\ignorespaces}
}

\newcommand\hz[1]{
    \ifthenelse{\equal{\showcomments}{yes}}{\textcolor{red}{\small [hz: #1~]}}{\ignorespaces}
}
\newcommand\yw[1]{
    \ifthenelse{\equal{\showcomments}{yes}}{\textcolor{purple}{\small [yw: #1~]}}{\ignorespaces}
}
\newcommand\yc[1]{
    \ifthenelse{\equal{\showcomments}{yes}}{\textcolor{green}{\small [yc: #1~]}}{\ignorespaces}
}
\newcommand\zz[1]{
    \ifthenelse{\equal{\showcomments}{yes}}{\textcolor{blue}{\small [zz: #1]}}{\ignorespaces}
}

\title{\Name: An Agentic Benchmark for LLM-Crafted Heuristics in Combinatorial Optimization}

\author{%
  Hongzheng Chen$^1$\thanks{Core Contributor}\enspace\quad
  Yingheng Wang$^1$\footnotemark[1]\enspace\quad
  Yaohui Cai$^1$\footnotemark[1]\enspace\quad
  Hins Hu$^1$\footnotemark[1]\enspace\quad
  Jiajie Li$^1$\footnotemark[1]\enspace\\
  \textbf{Shirley Huang$^2$\quad
  Chenhui Deng$^3$\quad Rongjian Liang$^3$\quad  Shufeng Kong$^1$}\\
  \textbf{Haoxing Ren$^3$\quad Samitha Samaranayake$^1$\quad Carla P. Gomes$^1$\quad Zhiru Zhang$^1$}\smallskip\\
  $^1$\ Cornell University\qquad
  $^2$\ Harvard University\qquad
  $^3$\ NVIDIA\smallskip\\
  \texttt{\{hzchen,yingheng\}@cs.cornell.edu}, \texttt{\{yc2632,zh223,jl4257\}@cornell.edu}
}

\iclrfinalcopy 
\begin{document}

\maketitle

\begin{abstract}
While Large Language Models (LLMs) have demonstrated significant advancements in reasoning and agent-based problem-solving, current evaluation methodologies fail to adequately assess their capabilities: existing benchmarks either rely on closed-ended questions prone to saturation and memorization, or subjective comparisons that lack consistency and rigor. In this work, we introduce \textbf{\Name}, an agentic framework designed for evaluating heuristic algorithms generated by LLMs for combinatorial optimization problems, characterized by clearly defined objectives and expansive solution spaces. \Name empowers LLMs to propose heuristics, receive evaluative feedback via code execution, and iteratively refine their solutions. We evaluate nine state-of-the-art models on various problems across domains such as computer systems, logistics, and biology, exposing persistent limitations in tool use, planning, and adaptive reasoning. To quantify performance, we propose the Quality-Yield Index (QYI), a metric that captures both solution pass rate and quality. Even top models like \GPTmini and \GeminiPro attain QYI scores of only 0.6, well below the expert baseline of 1. Our open-source benchmark aims to guide the development of LLMs toward more effective and realistic problem-solving in scientific and engineering domains.
\end{abstract}
\section{Introduction}
\label{sec:intro}

Recent advancements in Large Language Models (LLMs) have significantly expanded their capabilities in complex reasoning and agent-based problem-solving, enabling applications ranging from automated code generation~\citep{li2022alphacode,novikov2025alphaevolve} to dynamic decision-making systems~\citep{timo2023toolformer,yao2023react}.
Yet existing evaluation frameworks struggle to rigorously assess the full spectrum of these emergent abilities, often failing to capture the demands of real-world tasks that require iterative reasoning, creative algorithm design, and adaptive tool use. This shortcoming leaves a critical gap in understanding whether LLMs can move beyond pattern recognition to exhibit genuine problem-solving ingenuity in practice.


Current evaluation paradigms fall into two categories with distinct limitations.
\textbf{(1) Ground-truth-based objective benchmarks} rely on closed-form questions (e.g., multiple-choice mathematics problems) that have become susceptible to rapid performance saturation. Widely used benchmarks such as AIME~\citep{aime}, HumanEval~\citep{chen2021humaneval}, and GPQA Diamond~\citep{rein2024gpqa} now exhibit ceiling effects, with state-of-the-art models achieving over 80\% accuracy~\citep{openai_o3_o4,qwen3,gemini25}.
Even emerging evaluations like Humanity's Last Exam (HLE)~\citep{phan2025hle}, initially proposed as a rigorous PhD-level test, saw performance leap from 3\% to 25\% within months of release~\citep{openai_o3_o4}.
These benchmarks face a dual crisis: their static question banks risk data contamination as models ingest newer training data, while their closed-ended nature fails to reflect real-world problem-solving where solutions are neither unique nor predefined.
\textbf{(2) Judge-preference-based subjective evaluations}, such as Chatbot Arena~\citep{chiang2024chatbotarena}, take a different approach by assessing model quality through pairwise comparisons by humans or LLM-based proxies~\citep{zheng2023llmjudge}.
These benchmarks support a wide range of plausible outputs, making them better suited for open-ended tasks.
However, this flexibility introduces high variance: everyday communication tasks are inherently subjective, and judgments often prioritize superficial factors like response structure or emoji usage over substantive reasoning quality~\citep{singh2025leaderboardillusion,zhang2024emojibias}.
While recent efforts to automate evaluation with LLM-as-a-judge systems show promise, their reliability remains inconsistent across domains~\citep{krumdick2025llmjudgelimitation}, particularly for technical tasks requiring specialized expertise.

To address these limitations, we introduce \textbf{\Name}, a new evaluation paradigm with an agentic framework centered on combinatorial optimization problems, which naturally combine \emph{well-defined objectives} with \emph{large solution spaces}.
Rather than relying on well-known benchmarks such as SAT or TSP, we assess whether LLMs can produce high-quality solutions to novel yet foundational problems spanning computer systems~\citep{cai2025egraph,moffit2025iopddl}, scientific reasoning~\citep{chen2021automating,chen2016deep}, computational biology~\citep{dauparas2025protein,wijsman2012pedigree}, logistics~\citep{li2001metapdptw,graves1993cpp}, and electronic design automation~\citep{hofmann2025eqmap,cong2006sdc}.
They are ideal for benchmarking LLMs because they resist memorization due to their computational hardness, offer clear metrics for quantitative evaluation, and reflect real-world use cases where optimal solutions are tractable only for small instances.
Since no single heuristic dominates across all problems or instances~\citep{wolpert1997nfl}, the search space is rich and diverse. Tackling these challenges requires not only algorithmic knowledge but also heuristic reasoning, tradeoff navigation, and creative problem-solving---skills that are still underexplored in current LLM evaluations.
Our framework extends beyond conventional static evaluations by implementing an interactive agentic loop: LLMs generate heuristic algorithms, receive execution feedback from a code environment, and iteratively refine their solutions.
This process mirrors practical engineering workflows and enables deeper evaluation of multi-step reasoning, tool use, and instruction following.

Our benchmark systematically evaluates LLMs across four dimensions:
(1) \emph{tool-augmented reasoning} through integration with external libraries,
(2) \emph{multi-step planning} in decomposing complex problems into executable sub-tasks,
(3) \emph{instruction fidelity} in adhering to problem constraints, and
(4) \emph{iterative refinement} based on runtime feedback.
The framework uniquely probes practical creativity---the ability to adapt textbook algorithms or invent novel strategies for large-scale instances where exact methods like integer linear programming (ILP) may fail.

To capture both the number of feasible solutions and their quality relative to expert performance, we introduce a unified metric---the Quality-Yield Index (QYI)---which ranges from 0 (all outputs are incorrect or low-quality) to 1 (expert-level performance).
Empirical results reveal substantial performance gap: across nine diverse optimization problems, even state-of-the-art LLMs such as \GPTmini~\citep{openai_o3_o4} and \GeminiPro~\citep{gemini25} achieve QYI scores around 0.6, underscoring their limited effectiveness in realistic problem-solving settings.
These findings highlight the limitations of current benchmarks, which fail to capture the complex, real-world demands of computational problem-solving—where success requires integrating theoretical understanding, tool proficiency, and adaptive reasoning.
The contributions of this work are threefold:
\begin{itemize}
\item An open-source benchmark suite of nine combinatorial optimization problems that evaluates LLMs' multi-step reasoning capabilities through realistic programming tasks.
\item An end-to-end agentic framework supporting LLM solution generation, automated verification, quantitative evaluation with well-defined metrics, and iterative refinement. The resulting system can also serve as a testbed for exploring more advanced prompting techniques and evolutionary strategies.
\item A comprehensive empirical study of cutting-edge LLMs, uncovering their current limitations and offering actionable insights for the development of next-generation models and agents.
\end{itemize}
\section{Related Work}
\begin{table}[t]
\caption{Comparison with other recent benchmarks.}
\label{tab:comparison}
\centering
\resizebox{\linewidth}{!}{
\begin{tabular}{clcccc}\hline
\textbf{Subjects} & \textbf{Benchmark} & \textbf{\makecell[c]{Well-Defined\\Objective}} & \textbf{\makecell[c]{Large\\Solution Space}} & \textbf{\makecell[c]{Agentic\\Setting}} & \textbf{\makecell[c]{Evaluation\\Metrics}}\\\hline
\makecell[c]{Frontier\\Knowledge}
& \makecell[l]{Humanity's Last Exam\\(HLE)~\citep{phan2025hle}} & \rcmark & \xmark & \xmark & Accuracy\\\hline

\multirow{5}{*}{\makecell[c]{Software\\Engineering}}
& HumanEval~\citep{chen2021humaneval} & \rcmark & \xmark & \xmark & \textsc{pass@$k$}\\
& BigCodeBench~\citep{zhuo2025bigcodebench} & \rcmark & \xmark & \xmark & \textsc{pass@$k$}\\
& LiveCodeBench~\citep{jain2025livecodebench} & \rcmark & \xmark & \xmark & \textsc{pass@$1$}\\
& SWE-Bench~\citep{jimenez2024swebench} & \rcmark & \xmark & \xmark & \textsc{pass@$1$}\\
& Commit0~\citep{zhao2025commit0} & \rcmark & \xmark & \rcmark & Pass rate\\\hline

\makecell[c]{Performance\\Engineering}
& KernelBench~\citep{ouyang2025kernelbench} & \xmark & \rcmark & \xmark & \textsc{fast$_p$}\\\hline

\multirow{2}{*}{Daily-Life Tasks}
& Chatbot Arena~\citep{chiang2024chatbotarena} & \xmark & \rcmark& \xmark & ELO\\
& $\tau$-Bench~\citep{yao2025taubench} & \rcmark & \rcmark & \rcmark & \textsc{pass\textsuperscript{$\wedge$}$k$}\\\hline

\multirow{4}{*}{\makecell[c]{Combinatorial\\Optimization}}
& NPHardEval~\citep{fan2024nphardeval} & \rcmark & \xmark & \xmark & Accuracy\\
& GraphArena~\citep{tang2025grapharena} & \rcmark & \xmark & \xmark & Accuracy\\
& ALE-Bench~\citep{imajuku2025alebench} & \rcmark & \rcmark & \rcmark & ELO\\
& \textbf{\Name (This work)} & \rcmark & \rcmark & \rcmark & \textsc{solve$_s$@$i$}, QYI\\\hline
\end{tabular}
}
\end{table}

\textbf{LLMs for Combinatorial Optimization.}
Recent LLM-based combinatorial optimization (CO) methods follow two main paradigms. The first emphasizes formalization---translating natural language into structured optimization problems. This direction was initiated by the NL4Opt Competition~\citep{ramamonjison2023nl4opt}, with follow-up work improving domain-specific model training~\citep{xiao2023coe,jiang2025llmopt,li2025llmmilp} and prompting strategies~\citep{yang2024llmoptimizer,ali2024optimus,iklassov2024selfguiding}. While effective on benchmarks, these methods struggle to scale due to their reliance on exact solvers~\citep{gurobi}.
The second paradigm focuses on heuristic discovery. FunSearch~\citep{romera2024funsearch} and AlphaEvolve~\citep{novikov2025alphaevolve} use LLMs with evolutionary search to generate novel heuristics, but require evaluating thousands of candidates.
Recent approaches~\citep{ye2024reevo,liu2024eoh,dat2025hsevo} improve efficiency via metaheuristic templates, but still limit LLMs to filling in a small portion of the algorithm.
LLM4AD~\citep{liu2024llm4ad} offers a platform for evaluating such template-based methods.
In contrast, \Name removes reliance on templates or scaffolds. It tasks LLMs with generating complete, self-contained optimization programs, including custom data structures and end-to-end pipelines---better reflecting real-world CO challenges, where success depends on uncovering problem-specific structure and designing bespoke algorithms~\citep{wolpert1997nfl}.

\textbf{Evaluation on LLMs.}
As shown in Table~\ref{tab:comparison}, existing LLM benchmarks expose key limitations. Many focus on closed-ended tasks in domains like mathematics~\citep{aime}, programming~\citep{chen2021humaneval,zhuo2025bigcodebench,liu2023evalplus}, and specialized knowledge~\citep{rein2024gpqa,phan2025hle,dan2021math500}, with fixed ground-truths that are prone to data contamination (see \S~\ref{sec:intro}).
In contrast, open-ended benchmarks~\citep{chiang2024chatbotarena,ouyang2025kernelbench} encourage diverse outputs but often lack clear objectives, resulting in inconsistent evaluations.
Benchmarks like NPHardEval~\citep{fan2024nphardeval} and GraphArena~\citep{tang2025grapharena} assess exact solutions to small NP-hard instances, limiting real-world relevance where heuristic solutions are often preferred for scalability.
Our benchmark instead accepts any \emph{feasible} solution that satisfies constraints, enabling broader evaluation of algorithmic reasoning.
It tasks LLMs with synthesizing executable code, using external libraries, and refining solutions through execution feedback, mimicking realistic workflows.
ALE-Bench~\citep{imajuku2025alebench} and CO-Bench~\citep{Sun2025COBench} are concurrent efforts that focus on score optimization for classic CO problems using metaheuristics, whereas \Name targets practical, high-impact CO problems from scientific and engineering domains, requiring LLMs to discover problem-specific structure and design tailored heuristics.
Unlike ALE-Bench's ELO-style scoring, we propose fine-grained \solvesi\xspace and QYI metrics that reveal \emph{which stage} agents fail at and \emph{how close} they are to expert solutions in a multi-round reasoning setting, as detailed in \S~\ref{subsec:metric}.

\section{\Name: An Agentic Framework for Heuristic Generation}
\label{sec:framework}
In this section, we introduce our agentic framework for evaluating LLM reasoning via iterative heuristic generation, along with benchmark metrics for quantitative assessment.

\subsection{Overview}

As illustrated in Fig.~\ref{fig:overview}, our framework presents a formal problem description to the LLM, which is then prompted to generate a complete heuristic algorithm conforming to a standardized function signature.
It is subsequently compiled (for C++) or interpreted (for Python), and verified for correctness and performance.
Crucially, the framework incorporates a feedback loop: execution logs, verification outcomes, and evaluation costs from a small demonstration set are appended back to the prompt, enabling iterative refinement of the LLM-generated solution.

\subsubsection{Problem Description}
\label{subsub:problem_description}
\begin{figure}[t]
    \centering
    \includegraphics[width=0.9\linewidth]{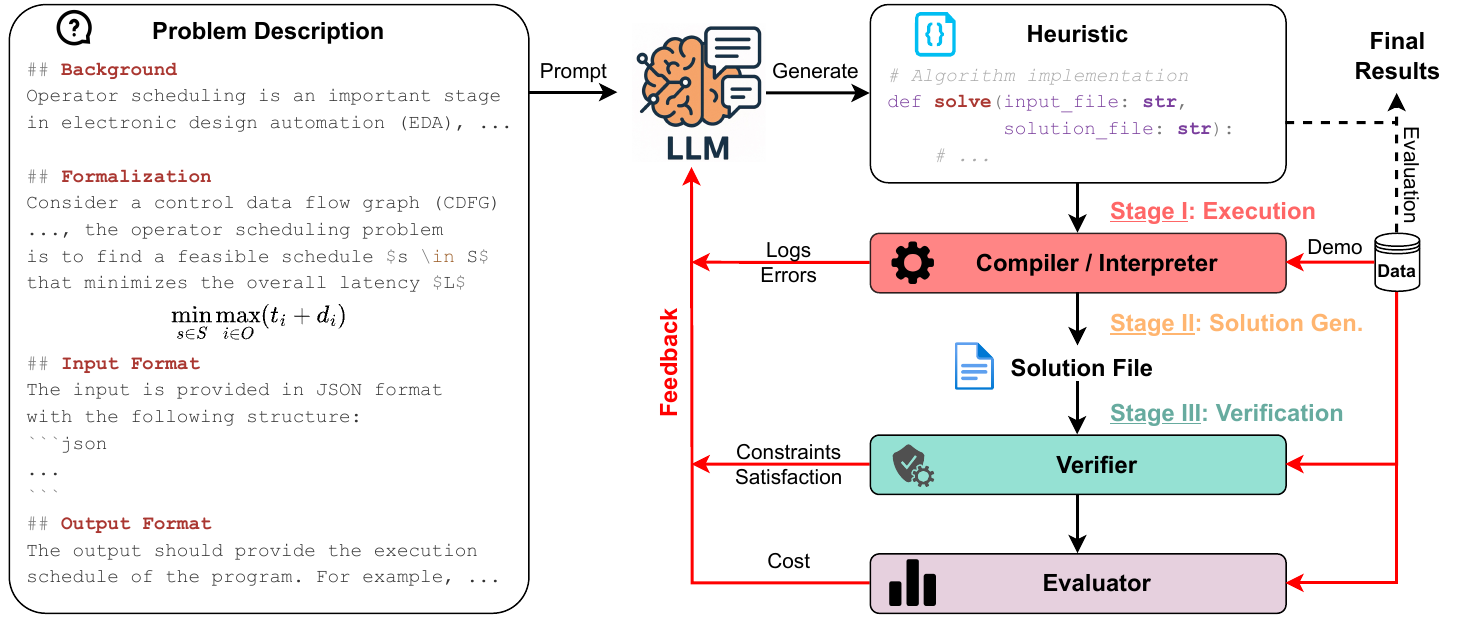}
    \caption{Overview of the \Name agentic framework for heuristic program generation, execution, and verification. We use operator scheduling as an example for the problem description.}
    \label{fig:overview}
\end{figure}

As shown on the left of Fig.~\ref{fig:overview}, we use operator scheduling~\citep{cong2006sdc}, a classic optimization problem in electronic design automation, as an example.
Each benchmark task is accompanied by a structured problem description with three main parts:
\textbf{(1) Background}: Introduces the optimization context and key terminology to help the LLM understand the problem setting.
\textbf{(2) Formalization}: Defines the optimization objective and constraints using mathematical notation (e.g., minimizing latency under hardware resource constraints), guiding the LLM toward objective-oriented algorithm design.
\textbf{(3) Input/Output Format}: Specifies the structure of input/output files, providing clear expectations for parsing and execution.
More details on the problem set can be found in \S~\ref{sec:benchmark}.

\subsubsection{Prompt Design}
Effective prompt engineering is crucial for leveraging LLMs' capabilities~\citep{wei2022cot,sahoo2024promptsurvey}.
We construct both system- and user-level prompts, tailored to each problem.
A complete prompt example is provided in Appendix~\ref{appendix:prompt}.

\textbf{System prompt.}
The system prompt includes machine configuration details (e.g., CPU cores, memory limits), available libraries with version numbers, and task-specific constraints such as execution timeouts.
This environment specification instructs the LLM to avoid relying on unrealistic assumptions or producing inefficient solutions that violate runtime limits.

\textbf{User prompt.}
In the initial iteration, the user prompt includes the problem description and a code skeleton with a predefined function signature.
As shown in Fig.~\ref{fig:overview}, the LLM is only provided the interface---function name, input path, and output path---without hints on data structures or algorithmic approache, contrasting with prior work~\citep{romera2024funsearch,liu2024eoh,ye2024reevo} that often handcrafts partial implementations or restricts the design space.
Here, LLMs must reason about the problem holistically: parsing inputs, constructing internal representations, and designing and implementing heuristics from scratch.

\subsubsection{Feedback Loop}
\label{subsubsec:feedback}
To emulate a few-shot in-context learning setup~\citep{dong2024incontext,liu2022gpt3icl,wu2023iclguide}, we partition the dataset into a small \emph{demonstration set} (around five instances) and a larger \emph{evaluation set}.
Demonstration data is used during the refinement loop to provide timely, example-based feedback to the LLM; the evaluation set is withheld until the model stabilizes its performance.

Each problem includes a domain-specific verifier and evaluator.
The verifier ensures constraint satisfaction (e.g., dependency preservation in operator scheduling), while the evaluator calculates the cost based on the given problem objective.
If the verifier fails, diagnostic messages are recorded.

After each iteration, we log the LLM-generated solution, execution trace, verification result, and the objective score.
These logs are appended to the prompt with the demonstration data in the next iteration, enabling the LLM to learn from past attempts and incrementally improve its output.

\subsection{Metric Design}
\label{subsec:metric}
Traditional LLM benchmarks predominantly rely on the \passk metric~\citep{chen2021humaneval,zhuo2025bigcodebench,jimenez2024swebench}, which measures the probability of generating a ground-truth solution within the top-$k$ samples.
While \passk is effective for single-turn tasks with deterministic ground truths, it falls short in capturing the iterative reasoning and problem-solving abilities required in our multi-round agentic setting.
Specifically, it does not reflect whether the LLM can understand problem constraints, debug based on feedback, or iteratively refine its solutions over multiple attempts.

To better evaluate LLMs in this complex setting, we introduce a new metric, denoted as \solvesi, which tracks the LLM's ability to solve constrained problems within $i$ iterations:
\[\solvesi:=\frac{1}{N}\sum_{n=1}^N\mathds{1}(\text{pass stage }s\text{ in the \emph{first} }i\text{-th iteration})\,,\]
where $N$ is the total number of instances that are fed to LLMs as inputs, and $s \in \{$\RNum{1}, \RNum{2}, \RNum{3}$\}$ denotes the pipeline stage that the solution must pass, each representing a key milestone in agentic reasoning:
\begin{itemize}
\item \textbf{Stage I: Execution}. The generated program must compile or interpret correctly with all necessary libraries included, and successfully perform basic I/O operations (e.g., reading and writing files).
\item \textbf{Stage II: Solution Generation}. The program must produce a non-empty output within the predefined timeout and adhere to the expected output format.
\item \textbf{Stage III: Verification}. The solution must satisfy all problem-specific constraints, as checked by a problem-specific verifier.
\end{itemize}

However, \solvesi\ only indicates whether a \emph{feasible} solution is eventually produced through the iterative process---it does not account for solution quality.
To address this, we additionally define separate metrics for quality and yield as follows:
\[
\textsc{Quality}=\frac{1}{\hat{N}}\sum_{n=1}^{\hat{N}} \min\left(1, \frac{c^\star_n}{c_n}\right)
\qquad
\textsc{Yield}=\frac{\hat{N}}{N}\,,
\]
where $c_n$ and $c^\star_n$ represent the cost (i.e., the optimization objective) of the LLM-generated and expert-provided solutions, respectively, and $\hat{N}$ is the number of instances that pass verification (Stage \RNum{3}) in one iteration.
We adopt the capped version of quality, which checks whether the LLM matches expert performance (up to a maximum of 1), though an uncapped version can also be used to measure cases where the LLM outperforms the expert.
We define a unified metric, the \emph{Quality-Yield Index (QYI)}, as the harmonic mean of quality and yield. This formulation, analogous to the F-score~\citep{van1979information}, penalizes imbalanced values more strongly than the arithmetic mean:
\[\textsc{QYI}=\frac{(1+\beta^2)\cdot \textsc{Quality}\cdot\textsc{Yield}}{(\beta^2\cdot\textsc{Quality}) + \textsc{Yield}}\,,\]
where $\beta$ controls the relative importance of \textsc{Yield} compared to \textsc{Quality}. By default, we use $\beta=1$ in our evaluation.
\textsc{QYI} captures both success rate and the relative quality of solutions, enabling holistic evaluation of an LLM's agentic reasoning capabilities, including its capacity for long-horizon planning and iterative refinement.
Additionally, we define a weighted QYI by averaging QYI scores with weights proportional to the number of instances in each task, to provide an aggregate measure of overall LLM performance. Nevertheless, we continue to report per-problem \textsc{Quality} and \textsc{Yield} to enable clearer inspection of tradeoffs across individual tasks.

\section{Benchmark Construction}
\label{sec:benchmark}

This section outlines the construction of our combinatorial optimization benchmark, detailing the principles behind problem selection and providing an overview of the resulting problem set.

\subsection{Problem Selection Criteria}
Our primary goal is to evaluate an LLM's capacity for reasoning rather than its ability to regurgitate well-known algorithms.
To this end, \emph{we intentionally exclude ubiquitous problems} such as TSP~\citep{robinson1949tsp} and SAT~\citep{schaefer1978sat}---problems that are so widely studied and frequently included in public datasets that they are likely memorized during pretraining.
Instead, we focus on problems that meet the following criteria:

\textbf{Limited exposure in the literature.}
For each candidate problem, we perform a Google Scholar search and retain it only if the most-cited paper has fewer than 1,000 citations (as of May 2025). This 1,000-citation cutoff is a practical criterion to exclude heavily standardized textbook CO problems that are almost certainly present in LLM training corpora, while still preserving well-defined, peer-reviewed problems. This \emph{empirical} threshold increases the likelihood that the LLM must genuinely reason and design new heuristics rather than rely on cached or widely publicized patterns.

\textbf{Clear natural-language specification with well-defined objectives.}
Each problem must be clearly expressible using plain language without the need for visual aids.
We encode mathematical objectives in \LaTeX\ to eliminate ambiguity, ensuring the LLM receives well-specified instructions.

\textbf{Large solution spaces.}
We focus on problems that admit vast solution spaces with many feasible outputs, encouraging creative exploration and reasoning rather than narrow pattern recognition~\citep{hughes2024openendedness}; in our benchmark, a single instance can present search spaces orders of magnitude larger than those of most existing benchmarks.

\textbf{Scalable data instances.}
Each problem includes two disjoint sets of instances: a small-scale demonstration set and a large-scale evaluation set, differing by at least an order of magnitude.
The demonstration set supports few-shot prompting and iterative refinement, while the evaluation set is reserved for final performance testing, as discussed in \S~\ref{subsubsec:feedback}.

\textbf{Reproducible expert baselines.}
Reference implementations are included in the benchmark repository to ensure fair comparisons in future studies.
These expert baselines are often problem-specific heuristic methods, representing the \emph{best-known results} from the literature, and are denoted as QYI 1.0 in our evaluation to highlight the performance gap. If some problems can be expressed as mixed-integer programs, we additionally provide formulations that can be run with commercial solvers like Gurobi. These are just for understanding the gap between heuristic solutions and the optimal solution on small-scale instances.

We prioritize domains with real-world impact, where even small gains yield significant societal or industrial benefits. Many selected problems remain open, with heuristics far from theoretical bounds---offering a compelling testbed for LLMs.

\subsection{Dataset Statistics}
\begin{table}[t]
\caption{Existing combinatorial optimization problems in our \Name benchmark.}
\label{tab:problems}
\centering
\resizebox{\linewidth}{!}{
\begin{tabular}{clccc}\hline
\textbf{Domain} & \textbf{Problem} & \textbf{References} & \textbf{Difficulty}\\\hline
\multirow{3}{*}{\makecell[c]{Electronic Design\\Automation (EDA)}}
& Operator scheduling & \cite{liu2024diffopsched,cong2006sdc} & $\bigstar$\\
& Technology mapping & \cite{hofmann2025eqmap,chen2004daomap} & $\bigstar\bigstar$\\
& Global routing & \cite{liang2024ispdcontest,liao2020deeprlrouting} & $\bigstar\bigstar\bigstar$\\\hline

\multirow{2}{*}{Compilers}
& E-graph extraction & \cite{cai2025egraph,willsey2021egg} & $\bigstar$\\
& Intra-operator parallelism & \cite{moffit2025iopddl,zheng2022alpa} & $\bigstar\bigstar$\\\hline

\multirow{2}{*}{\makecell{Computational \\ Biology}}
& Protein sequence design & \cite{dauparas2025protein,kleinberg1999protein} & $\bigstar$\\
& Mendelian error detection & \cite{lundgren2025pedigree,sanchez2008mendelian} & $\bigstar\bigstar$\\\hline

\multirow{2}{*}{Logistics}
& Airline crew pairing & \cite{korte2025aircraft,aggarwal2023crewpairing} & $\bigstar\bigstar$\\
& Pickup and delivery w/ time windows & \cite{taniguchi2025timewindow,li2001metapdptw} & $\bigstar\bigstar\bigstar$\\\hline
\end{tabular}
}
\end{table}


The initial release of \Name spans nine optimization problems across scientific and engineering domains, covering fundamentally different types such as covering, scheduling, and routing in Table~\ref{tab:problems}.
Each problem provides five demonstration instances that capture different constraint requirements and provide representative guidance, and dozens of large-scale evaluation instances, totaling 218 (distribution in Appendix~\ref {appendix:dataset}).
A small number of demonstration instances is a practical consideration to reduce evaluation time per iteration and also aligns with few-shot LLM usage~\citep{wei2022cot,dong2024incontext}. The effectiveness of the demonstration set is further analyzed in \S~\ref{subsec:ablation}.

For \emph{each} instance, the search space is enormously large, growing combinatorially with problem size. This yields many distinct valid solutions rather than a single ground truth, making exhaustive search entirely infeasible. In several tasks, the search space far exceeds even astronomical scales (e.g., $10^{65{,}000}$ for intra-operator parallelism), and when combined with hard constraints and non-linear objectives, the resulting problems cannot be handled by commercial solvers and are substantially more challenging than typical closed-form optimization benchmarks.
All datasets are derived from realistic sources and real-world applications, enhancing the benchmark's practical relevance.
Notably, most problems are NP-hard and feature complex constraints, resulting in a compact yet highly challenging problem suite.
In addition, we reserve hundreds of instances as private test sets for future release.
A detailed description of each problem is provided in Appendix~\ref{appendix:problem_set}, and empirical results (\S~\ref{sec:experiment}) show that the benchmark remains substantially difficult for state-of-the-art LLMs.

To ensure clarity and correctness, we adopt a human-in-the-loop process for problem specification.
A human annotator first drafts the natural-language description, then uses a weaker LLM such as DeepSeek-V3~\citep{deepseek_v3} to highlight any unclear or ambiguous statements. The human and model iteratively resolve discrepancies until the specification is unambiguous and fully aligned with the intended semantics. Importantly, this procedure is solely for refining the problem descriptions but \emph{not} for improving solver performance.
The assumption is that if a weaker model can successfully validate the specification's unambiguity and coherence, a stronger model will inherently possess sufficient understanding.
The full prompt template used for refining problem descriptions is provided in Appendix~\ref{appendix:refinement_problem}.

Each problem includes a task-specific verifier and evaluator to assess solution pass rate and quality. A separate reviewer ensures the expert solver reproduces published results and passes both checks.

Looking forward, we plan to extend \Name\ along two axes: (1) \emph{breadth}, by incorporating additional combinatorial optimization problems from underexplored scientific domains;
and (2) \emph{depth}, by scaling existing problems to larger instance sizes and tighter constraint settings.
Community contributions are welcome, provided new problems satisfy the selection criteria articulated above.
\section{Evaluation}
\label{sec:experiment}
To evaluate the reasoning capabilities of LLMs on CO problems, we benchmark nine prominent models released in late 2024 and mid-2025 (\S~\ref{appendix:models}).
These models represent the current state-of-the-art in general-purpose LLMs and rank among the top entries on OpenRouter~\citep{openrouter} and Chatbot Arena leaderboards~\citep{chiang2024chatbotarena}.
We exclude smaller models due to the complexity of the benchmark tasks.
All evaluations are conducted via official APIs to ensure reproducibility. We adopt the agentic workflow in Fig.~\ref{fig:overview}, constraining each model to generate Python programs that solve the given problems under fixed resource limits: a maximum of 8 CPU cores and problem-specific timeouts.
We also allow the models to access the given Python libraries for external tool use.
Full details of the experimental settings and results of each problem can be found in \S~\ref{appendix:exp}.
Notice the main goal of these experiments is \emph{not} to present an optimal pipeline, but rather to establish a baseline for current LLM capabilities and to demonstrate that \Name serves as a common testbed on top of which more advanced prompting strategies and sophisticated agentic workflows can be developed. We further discuss these potential improvements in \S~\ref{sec:discussion}.

\subsection{Overall Performance}
\begin{table}[t]
\caption{Overall \solvesi\ metric of models on the whole \Name benchmark.}
\label{tab:overall}
\centering
\resizebox{\linewidth}{!}{
\begin{tabular}{l|>{\columncolor{gray!20}}ccc|ccc|ccc}\hline
 &\multicolumn{3}{c|}{\textsc{solve$_\text{\RNum{3}}$}}
 &\multicolumn{3}{c|}{\textsc{solve$_\text{\RNum{2}}$}}
 &\multicolumn{3}{c}{\textsc{solve$_\text{\RNum{1}}$}}\\
\textbf{Model} & @10 & @5 & @1 & @10 & @5 & @1 & @10 & @5 & @1\\\hline
\DeepSeekV & 46.8\% & 42.7\% & 14.2\% & 87.6\% & 83.0\% & 66.1\% & \textbf{100.0\%} & \textbf{100.0\%} & 90.8\% \\
\DeepSeekR & 73.4\% & \textbf{72.9\%} & 44.0\% & 88.1\% & 88.1\% & 60.6\% & \textbf{100.0\%} & \textbf{100.0\%} & 71.6\% \\
\GeminiFlash & 67.4\% & 58.3\% & 25.2\% & 83.9\% & 79.4\% & 56.4\% & \textbf{100.0\%} & \textbf{100.0\%} & 72.9\% \\
\GeminiPro & 65.1\% & 64.2\% & 20.2\% & 89.4\% & 89.0\% & 42.7\% & \textbf{100.0\%} & \textbf{100.0\%} & 51.4\% \\
\Llamal & 35.8\% & 33.5\% & 6.0\% & 84.9\% & 74.3\% & 8.3\% & 85.3\% & 85.3\% & 13.3\% \\
\Llamas & 33.9\% & 33.9\% & 20.6\% & 78.4\% & 78.4\% & 40.4\% & 99.5\% & 99.5\% & 61.9\% \\
\Qwen & 45.9\% & 45.4\% & 38.5\% & 86.2\% & 83.0\% & 56.0\% & \textbf{100.0\%} & \textbf{100.0\%} & 70.6\% \\
\Claude & 60.1\% & 58.7\% & 9.2\% & 97.7\% & 97.7\% & 41.3\% & \textbf{100.0\%} & \textbf{100.0\%} & 60.1\% \\
\GPTmini & \textbf{74.8\%} & 69.7\% & \textbf{53.2\%} & \textbf{100.0\%} & \textbf{100.0\%} & \textbf{93.1\%} & \textbf{100.0\%} & \textbf{100.0\%} & \textbf{100.0\%} \\
\hline
\end{tabular}
}
\end{table}

\begin{figure}[t]
\centering
\begin{minipage}{0.49\linewidth}
  \centering
  \includegraphics[width=\linewidth]{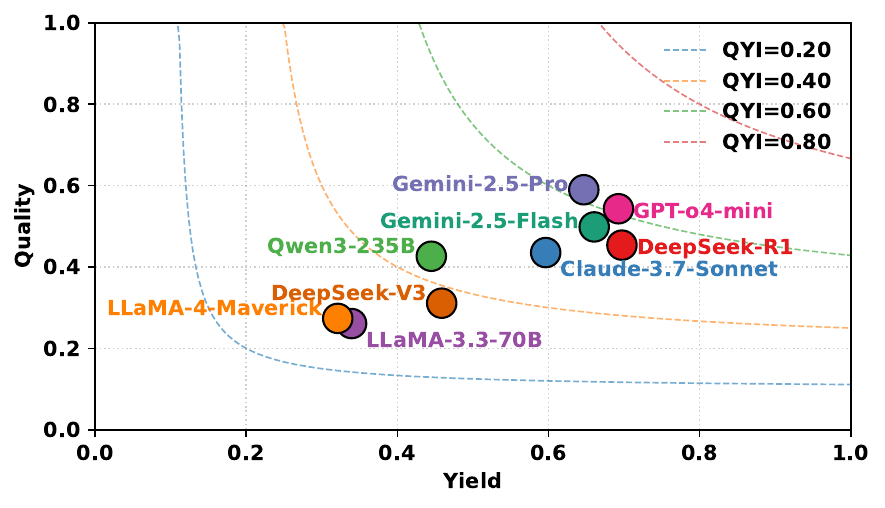}
\end{minipage}
\hfill
\begin{minipage}{0.49\linewidth}
  \centering
  \includegraphics[width=\linewidth]{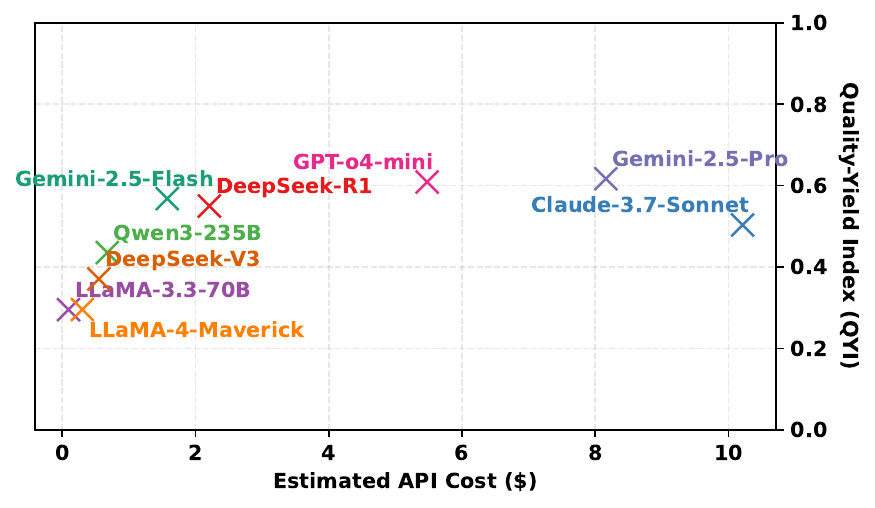}
\end{minipage}
\caption{Quality-Yield Index and estimated API cost of different models.}
\label{fig:qyi}
\end{figure}

For the overall evaluation, we fix the generation temperature at 0, following standard practice in recent LLM benchmarks~\citep{ouyang2025kernelbench,yao2025taubench,phan2025hle}.
This ensures deterministic outputs and eliminates randomness across runs.
Notably, OpenAI's o-series models only support a fixed temperature of 1.0~\citep{openai_o3_o4}.
We measure the multi-round performance using the \solvesi\ metric, where $i\in\{1,5,10\}$ indicates the number of iterations allowed.

As shown in Table~\ref{tab:overall}, most LLMs fail to solve a large fraction of test cases within a single attempt, as reflected in the \solve{3}{1} score. Increasing the number of iterations generally improves performance across all models. For instance, the \textsc{solve}$_{\textsc{\RNum{3}}}$ success rate rises from 53.2\% to 74.8\% for \GPTmini as $i$ increases, underscoring the importance of iterative refinement in improving LLM-generated solutions. Among all models, \GPTmini and \DeepSeekR demonstrate high success rates across multiple iterations, highlighting their stronger program repair capabilities.

To assess solution quality, we compare the final LLM-generated programs to expert-designed solutions using the weighted QYI metric defined in \S~\ref{subsec:metric}. As illustrated in Fig.~\ref{fig:qyi}, a substantial performance gap remains: even the best-performing model, \GeminiPro, achieves a QYI of only 0.62, indicating that its solutions are, on average, just 60\% as effective as expert-crafted ones.
Several models, such as \Llamas and \Llamal, produce results with QYI scores below 30\%, highlighting their limited effectiveness on these tasks. We also estimate the API cost for each model and find that \GeminiFlash offers the best cost-efficiency relative to its achieved QYI.

We further compare state-of-the-art open-source evolutionary frameworks under the same setting.
We fix the outermost evolutionary loop to 10 iterations and use a population size of 10; in total, each method therefore produces hundreds of candidate programs due to multi-step reasoning within each iteration. Further increasing the population size leads to context-length overflows for most of the problems inside \Name.
All frameworks use \GeminiPro, the best-performing model in our benchmark, as the base LLM. The initial candidate program is generated directly from the problem description.
As shown in Table~\ref{tab:evo}, these frameworks perform poorly, often worse than the baseline model. Their main weakness is the lack of incorporating program execution feedback (errors and verification results, \S~\ref{subsubsec:feedback}), and their search process breaks context across iterations, causing the system to repeatedly patch the same flawed initial program without real progress.
This highlights both the complexity of our benchmark and the need for LLMs to reason more deeply about problem-specific strategies. These frameworks originally only target toy problems under 20 lines of code (e.g., TSP, bin packing), while our benchmark typically requires 300+ lines, making strategy discovery essential rather than relying on prebuilt metaheuristics.
These results also underscore the importance of better prompt design and context engineering. Simply appending all sampled programs to the prompt does not scale to our benchmark, limiting the population size per iteration. Moreover, improved mechanisms are needed to consolidate and reconcile feedback from different sampled candidates so that refinements do not conflict with one another.

\begin{table}[htbp]
\begin{minipage}{0.6\linewidth}
\centering
\captionof{table}{Performance of evolutionary frameworks.}
\label{tab:evo}
\small
\begin{tabular}{ccc}\hline
\textbf{Frameworks} & \textbf{\solve{3}{10}} & \textbf{QYI}\\\hline
\GeminiPro & 0.6514 & 0.6170\\\hline
\makecell{HSEvo\\\citep{dat2025hsevo}} & 0.5000 & 0.4491 \\\hline
\makecell{ReEvo\\\citep{ye2024reevo}} & 0.4771 & 0.4486 \\\hline
\makecell{EoH\\\citep{liu2024eoh}} & 0.4954 & 0.4492\\\hline
\end{tabular}
\end{minipage}
\begin{minipage}{0.39\linewidth}
\centering
\captionof{table}{Ablation study on pickup and delivery with time windows.}
\label{tab:ablation}
\small
\begin{tabular}{c|c}\hline
\makecell{\textbf{\# of Demos /}\\\textbf{\# of Feedback Rounds}} & \textbf{QYI}\\\hline
5/10 & 0.4196\\\hline
3/10 & 0.2829\\\hline
0/10 & 0.2351\\\hline
5/5 & 0.3330\\\hline
5/1 & 0.2350\\\hline
\end{tabular}
\end{minipage}
\end{table}

\begin{figure}[t]
\centering
\begin{minipage}{0.495\textwidth}
    \centering
    \includegraphics[width=\linewidth]{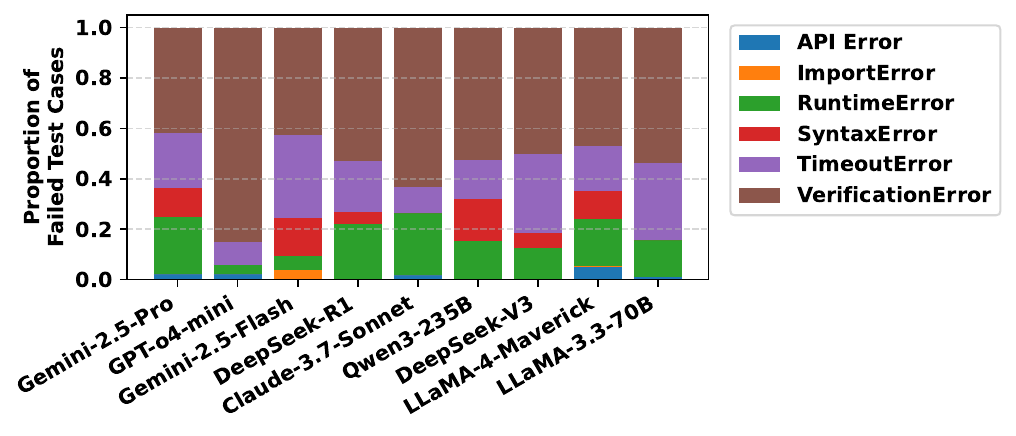}
    \caption{Error classifications.}
    \label{fig:error}
\end{minipage}
\hfill
\begin{minipage}{0.495\textwidth}
\centering
    \includegraphics[width=\linewidth]{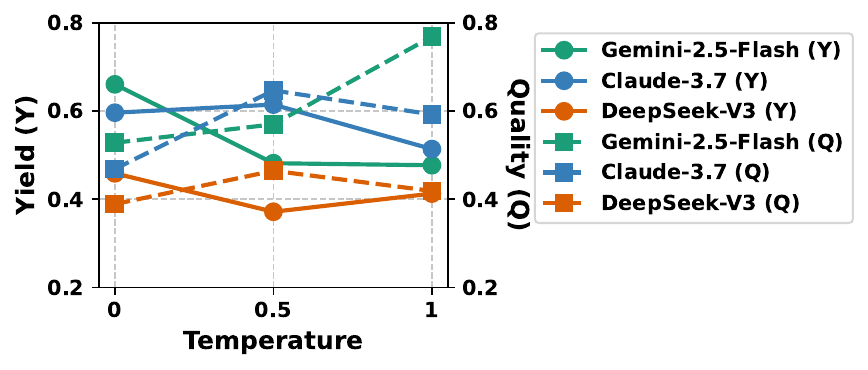}
    \caption{Quality-Yield tradeoff.}
    \label{fig:temperature}
\end{minipage}
\end{figure}

To identify common failure modes, we analyze and categorize the most common error types produced by the evaluated models, as shown in Fig.~\ref{fig:error}. These include:
(1) Hallucinated APIs: using nonexistent or outdated library calls.
(2) Incorrect algorithmic logic: flawed implementation even when the general approach is reasonable.
(3) Constraint misunderstanding: ignoring or misinterpreting problem constraints.
(4) Timeouts: no output or the execution time exceeds the given constraints.
Additional error cases and examples are listed in Appendix~\ref{appendix:errors}.

\subsection{Ablation Study}
\label{subsec:ablation}
To assess the robustness and sensitivity of LLM performance under different settings, we conduct a set of ablation experiments with full details in Appendix~\ref{appendix:exp}.

\textbf{Temperature.} We evaluate three representative models across the QYI spectrum using temperatures $T \in \{0.0, 0.5, 1.0\}$. Fig.~\ref{fig:temperature} shows that higher $T$ increases diversity and quality but lowers yield due to more invalid outputs (\S~\ref{appendix:temperature}). Greedy decoding ($T=0$) has maximum yield with suboptimal quality, while stochastic sampling ($T=1$) achieves better quality at the cost of solving fewer problems. This reveals a fundamental trade-off between quality and yield that future LLMs must address.

\textbf{Few-shot demonstrations.}  We assess the impact of in-context examples by comparing zero-shot, half-shot, and full-shot prompts. Due to budget constraints, these experiments are conducted on a few representative models. Specifically, we evaluate \GeminiPro on the pickup and delivery problem---one of the most challenging tasks in our benchmark (full results in \S~\ref{appendix:few-shot}). As shown in Table~\ref{tab:ablation}, providing more informative demonstrations significantly boosts the overall performance, especially for tasks involving unfamiliar domains or requiring long-horizon reasoning.

\textbf{Feedback rounds.} To evaluate the role of iterative refinement, we vary the number of feedback rounds given to LLMs (1, 5, and 10), keeping the temperature fixed at 0. The results in Table~\ref{tab:ablation} show that later iterations frequently fix logic errors or constraint violations from earlier attempts, underscoring the value of multi-round reasoning. We provide further analysis in \S~\ref{subsec:case_study} and \S~\ref{appendix:feedback-rounds}.


\begin{figure}[t]
    \centering
    \includegraphics[width=\linewidth]{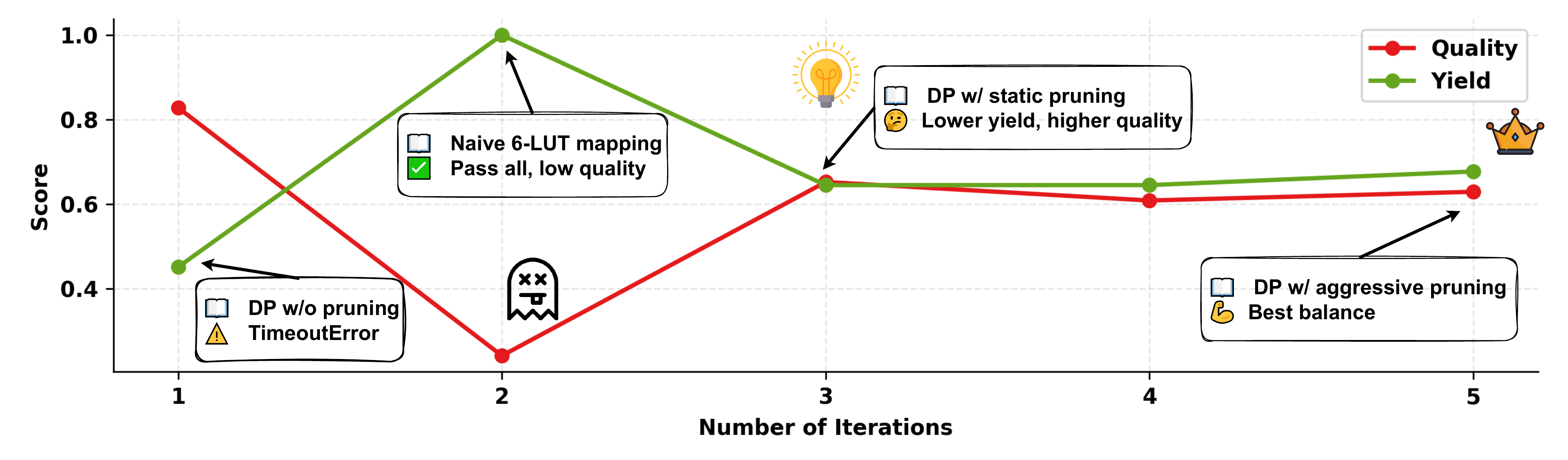}
    \caption{One iterative example of \GPTmini on the technology mapping problem.}
    \label{fig:case_study}
\end{figure}

\subsection{Case Study}
\label{subsec:case_study}
We present a case study on technology mapping~\citep{chen2004daomap} to highlight both the promise and current limitations of LLMs, where the task is to cover a logic network with $K$-input lookup tables (LUTs) while minimizing their total count; we fix $K=6$.
As an expert baseline, we use ABC~\citep{abc_tool}, a state-of-the-art logic synthesis tool that leverages optimized cut enumeration and dynamic programming (DP)-based covering. We find that top-performing LLMs, such as \GPTmini and \GeminiPro, can mimic similar heuristic strategies and iteratively refine them through feedback. Fig.~\ref{fig:case_study} shows \GPTmini explores a range of approaches over multiple iterations, evolving from naive mappings to sophisticated DP-based heuristics with pruning. It finally converges on a strategy that effectively balances yield and quality, achieving the highest QYI.
The full generated programs across those iterations are listed in \S~\ref{appendix:case-study}.

Nonetheless, a substantial gap remains between LLMs and expert tools ($\sim$60\% of expert performance), due to the latter's extensive use of domain-specific optimizations and efficient implementations. This suggests that while LLMs can learn and refine heuristic algorithms, they are not yet capable of generating solutions with expert-level performance in real-world complex optimization tasks.

\section{Discussion and Limitation}
\label{sec:discussion}

To guide future LLM development, we identify two key challenges: \textbf{(1) Correctness}---reducing hallucinations (e.g., incorrect API usage) and improving instruction-following to strictly satisfy problem constraints; and \textbf{(2) Performance}---navigating enormous search spaces to discover high-quality solutions. Potential enhancements such as longer context windows and advanced search strategies like iterative best-of-N (\S~\ref{appendix:bestofn}) may help address these challenges.

While our benchmark and framework offer a promising foundation for evaluating LLMs on combinatorial optimization problems, several limitations remain that suggest directions for future work.

First, all experiments are run in Python, which eases adoption but incurs execution overheads. We report preliminary C++ results in \S~\ref{appendix:cpp}, but full integration remains challenging due to reliance on domain-specific libraries and the difficulty of generating efficient, correct, and parallel C++ code.

Second, our current agentic pipeline only uses a standard configuration as a baseline and does not incorporate advanced prompting or multi-agent strategies. There are several promising directions for improving context efficiency, such as using summarization~\citep{sahoo2024promptsurvey} or prompt compression~\citep{chuang2024promptcompress}. Automatic prompt engineering further expands the design space, with approaches based on gradient descent~\citep{pryzant2023protegi} or genetic algorithms~\citep{guo2024evoprompt} for searching optimal prompts. In addition, multi-agent designs in which different agents handle different components of a decomposed problem have shown strong potential for tackling more complex tasks~\citep{zhou2025multiagent,gottweis2025aicoscientist}.

Third, the iterative self-refinement process in our agentic workflow can be interpreted as a form of test-time scaling (TTS), analogous to compute-optimal scaling strategies~\citep{snell2024scaling}. This perspective creates opportunities to incorporate techniques such as best-of-N sampling~\citep{stiennon2020learningsummarize}, beam search~\citep{xie2023beamsearch}, evolutionary algorithms~\citep{novikov2025alphaevolve,ye2024reevo}, and external autotuner~\citep{van2025llameahpo}, especially with increased iteration budgets. Furthermore, with a robust verifier in place, our framework provides a natural platform to investigate self-verification capabilities~\citep{kumar2024training,zhang2025and,zheng2024processbench}, a promising avenue toward greater LLM autonomy.

Finally, our evaluation pipeline currently relies on formally defined and computationally efficient proxy metrics. While these metrics are useful for benchmarking, they may not fully reflect real-world performance. This gap is especially apparent in (1) scientific domains, where solution quality must ultimately be validated through physical experiments, and (2) engineering domains like EDA, where quality must be confirmed through time-consuming backend synthesis. Bridging the gap---while managing the latency introduced by longer feedback loops---remains a key challenge for future work.


We believe \Name can serve as a shared testbed and foster interdisciplinary collaboration.



\section*{Ethics Statement}
In the development of this work, we have been guided by the ICLR Code of Ethics, aiming to contribute positively to the field of combinatorial optimization and human well-being.
We acknowledge that, like any new dataset, the content presented could be misused.
Our intention for this work is to advance the field of combinatorial optimization for societal benefit, and we encourage the research community to consider and mitigate potential negative consequences in future applications.

We recognize that the large language models utilized may reflect existing biases.
By primarily using publicly available models, we aim to be transparent and have avoided using private user data.
The use of LLMs is listed in Appendix~\ref{appendix:llmuse}.
We advocate for continued research into identifying and mitigating biases in LLM-crafted heuristics to ensure fairness and prevent discrimination.

\section*{Reproducibility Statement}
Our framework, dataset, and benchmark are released under an open-source license and are available in the supplementary material during the review process. All models are accessible via the public API through OpenRouter~\cite{openrouter}. The models included in our benchmark are listed in Appendix~\ref{appendix:models}, with detailed experimental settings provided in Appendix~\ref{appendix:exp}. To support reproducibility, we also provide step-by-step instructions in our repository.

\newpage
\bibliography{iclr2026_conference}
\bibliographystyle{iclr2026_conference}

\clearpage
\newpage
\appendix
\section*{Appendix} 
\addcontentsline{toc}{section}{Appendix}
\startcontents[appendix]  
\printcontents[appendix]{l}{1}{\setcounter{tocdepth}{2}}
\newpage
\section{The Use of Large Language Models (LLMs)}
\label{appendix:llmuse}
In this work, we design an agentic benchmark and evaluate the performance of various LLMs on combinatorial optimization problems. Model specifications are provided in Appendix~\ref{appendix:models}, and experimental details are described in Appendix~\ref{appendix:exp}. We ensure that all uses of LLMs are transparent, and our results can be fully reproduced from the submitted supplementary material. Beyond their use in experiments, we also employ LLMs to polish the writing. Importantly, we do \emph{not} use LLMs to propose ideas or design experiments.
\section{Prompt Design}
\label{appendix:prompt}
In this section, we detail the system and user prompts used by the LLM agent, as well as the auxiliary prompt employed to enhance our problem descriptions.

\subsection{System Prompt}
Each iteration of our benchmark begins with a task-agnostic system prompt that instructs the LLM to generate and iteratively refine executable heuristics for combinatorial optimization problems. This system prompt is followed by a task-specific problem statement and an input/output specification. The prompt includes placeholders -- highlighted in red -- that are dynamically instantiated at runtime for each task. For instance, {\color{red}{\{NUM\_CPU\_CORES\}}} represents the CPU core limit for the task (default: 8), and {\color{red}{\{TIMEOUT\}}} specifies the wall-clock time limit (default: 10 seconds).

\begin{tcolorbox}[title=System Prompt, colframe=black!85, colback=white, fonttitle=\bfseries]
You are a world-class optimization expert and algorithmic problem solver. Your task is to develop a highly efficient solution to the following optimization problem. Please analyze the problem background, mathematical formulation, and I/O specifications with extreme rigor and attention to detail.

\bigskip
Your mission is to devise and implement the most performant algorithm possible, optimizing for both computational efficiency and solution quality. You should leverage your deep knowledge of algorithms, data structures, and optimization techniques to craft a powerful solution. You have complete freedom in your algorithmic approach. Think systematically and creatively. Your goal is to push the boundaries of what's possible within the computational constraints. Please strictly follow the instructions below.
\begin{enumerate}
    \item A problem template is provided below. You only need to implement the solve function. Do NOT modify the function signature including the data types of the input arguments. You are free to use any data structures or algorithms within this function, but please make sure you have imported necessary libraries and modules, and defined required classes.
    \item The evaluation machine has {\color{red}\{NUM\_CPU\_CORES\}} CPU cores and sufficient memory to run your program. The time limit for this question is {\color{red}{\{TIMEOUT\}}} seconds. You are free to implement parallel algorithms where appropriate to maximize performance.
    \item The Python version is 3.12. You may use any standard Python libraries and only the following third-party libraries:
    \begin{itemize}
        \item numpy==2.2.5
        \item networkx==3.4.2
        \item pandas==2.2.3
    \end{itemize}
    \item Your response should consist of a complete implementation of the `solve' function. Do NOT include any explanations, comments, additional text, or Markdown formatting.
    \item You will receive execution feedback after the user runs your program, including runtime metrics and correctness evaluation.
\end{enumerate}
\end{tcolorbox}
\bigskip

\subsection{User Prompt}
For each problem, the first iteration begins with the following user prompt, which introduces the task and its objective to the LLM, along with a program template that the model is expected to complete.

\begin{tcolorbox}[title=User Prompt, colframe=black!85, colback=white, fonttitle=\bfseries]
\# Problem Information \newline
\textcolor{red}{\{PROBLEM DESCRIPTION\}}\newline

\# Program Template
\begin{minted}[fontsize=\footnotesize]{python}
def solve(input_file: str, solution_file: str):
    """
    Solve the optimization problem.

    Please do NOT change the function name and arguments.
    Inputs should be read from input_file
    and outputs should be written to solution_file.
    Input and output formats have been specified in the
    problem statement.
    """
    raise NotImplementedError(
        "This is a placeholder implementation you need to fill in."
    )
\end{minted}
\end{tcolorbox}

\subsection{Prompts for Improvement Guidance}
Based on the feasibility of the final outputs, we issue one of two improvement prompts in subsequent iterations.
If any test cases fail, we provide the following prompt:

\begin{tcolorbox}[title=Improvement Guidance Case 1, colframe=black!85, colback=white, fonttitle=\bfseries]
\# Feedback from Previous Iteration (Iteration \texttt{\{iteration-1\}}) \newline
These are the test cases and results from the previous iteration:

\#\# Test Case 1: \texttt{\{test\_name\}} \newline
**Input File:** \newline
\{content\} \newline
**Result:**\newline
\texttt{\{execution\_message\}} \newline

\#\# Test Case 2: \texttt{\{test\_name\}} \newline
**Input File:** \newline
\{content\} \newline
**Result:**\newline
\texttt{\{execution\_message\}} \newline

...\newline

\# Improvement Guidance \newline
The program failed to produce valid solutions for some test cases. Please fix the following issues:
\begin{enumerate}
    \item Check for compilation errors or runtime exceptions.
    \item Ensure the program handles all edge cases and meets the problem constraints correctly.
    \item Verify that the input and output format match the expected format.
    \item Make sure all required functions are implemented correctly, and no external forbidden libraries are used.
    \item If the program is not able to produce valid solutions for any test case, please try to find the root cause and fix it.
    \item If the program is able to produce valid solutions for some test cases, please try to improve the solution.
\end{enumerate}
\end{tcolorbox}

Otherwise, if all test cases pass verification, we issue the following prompt:

\begin{tcolorbox}[title=Improvement Guidance Case 2, colframe=black!85, colback=white, fonttitle=\bfseries]
\# Feedback from Previous Iteration (Iteration \texttt{\{iteration-1\}}) \newline
...\newline

\# Improvement Guidance \newline
Please carefully observe the problem structure and improve upon this program by:
\begin{enumerate}
    \item Addressing any weaknesses in the previous approach.
    \item Introducing more advanced or efficient algorithms.
    \item Focusing on improving performance for test cases.
\end{enumerate}
Your goal is to improve the solution for as many test cases as possible, with special attention to those where the previous solution performed poorly.
\end{tcolorbox}

\subsection{Refinement Prompt for Problem Descriptions}
\label{appendix:refinement_problem}
To ensure clarity and correctness in problem specification, we employ a human-in-the-loop process. Specifically, we prompt a weaker LLM to flag any unclear or ambiguous statements in the task description. The following prompt is used for this purpose:

\begin{tcolorbox}[title=Refinement Prompt for Problem Descriptions, colframe=black!85, colback=white, fonttitle=\bfseries]
If you were to solve the programming task below, do you have any questions? Is there anything I should clarify before you begin writing code?\newline

\# Problem Description\newline
\textcolor{red}{\{PROBLEM DESCRIPTION\}}
\end{tcolorbox}

\subsection{Example Problem Description}
The following provides an example problem description for operator scheduling. For other problems, please refer to our repository.

\begin{minted}[breaklines=true, breakanywhere=true, frame=single,fontsize=\scriptsize]{text}
## Background

High-level synthesis (HLS) is an important stage in electronic design automation (EDA), aimed at translating a high-level program specification (e.g., written in C/C++ or SystemC) into a cycle-accurate hardware implementation. After the program is parsed and analyzed, it is typically transformed into an intermediate representation known as a Control Data Flow Graph (CDFG). This graph captures the operations (e.g., arithmetic, memory accesses) and their control/data dependencies. The CDFG can further be processed into a Directed Acyclic Graph (DAG) to facilitate scheduling and optimization.

One of the core challenges in HLS is operator scheduling, which determines the exact control step (or cycle) at which each operation is executed, while satisfying data dependencies and resource constraints. Efficient scheduling plays a critical role in optimizing design quality in terms of performance, area, and power.

## Formalization

Consider a CDFG with $n$ operation nodes $o_i$, where $i \in O = \{1, 2, \ldots, n\}$, and a precedence relation $\prec$ on $O$ that captures operation dependencies. Each operation $o_i$ is associated with a cycle delay $d_i \in \mathbb{Z}^+$ and a resource type $r_i \in R = \{1, 2, \ldots, k\}$. Let $T = \{0, 1, 2, \ldots, L\}$ represent the set of control steps (c-steps), and define a schedule as an $n$-tuple $s = (t_1, t_2, \ldots, t_n)$, where $t_i \in T$ denotes the start time (c-step) of operation $o_i$.

A schedule $s$ is feasible if it satisfies all data dependencies:  
$\forall i, j \in O: i \prec j \Rightarrow t_i + d_i \leq t_j$.  
Let $S$ denote the set of all feasible schedules. For a given schedule $s$, let $N_r(t)$ be the number of operations that use resource $r$ in control step $t$, and define the total usage of resource $r$ as $N_r = \sum_{t \in T} N_r(t)$.

Given a bound $G_r$ on the number of available instances for each resource type $r \in R$, the operator scheduling problem is to find a feasible schedule $s \in S$ that minimizes the overall latency $L$, defined as  
$\min_{s \in S} \max_{i \in O} (t_i + d_i)$,  
subject to the resource constraints  
$\forall r \in R, t \in T: N_r(t) \leq G_r$.

## Input Format
The input is provided in JSON format with the following structure:

```json
{
  "name": "input",
  "delay": {
    "mul": 3,
    "sub": 1
  },
  "resource": {
    "mul": 2,
    "sub": 1
  },
  "nodes": [
    ["n1", "mul"],
    ["n2", "mul"],
    ["n3", "sub"]
  ],
  "edges": [
    ["n1", "n3", "lhs"],
    ["n2", "n3", "rhs"]
  ]
}
```

Where:
- `name`: Name of the input graph
- `delay`: Maps each resource type to its execution delay in cycles
- `resource`: Maps each resource type to the number of available functional units
- `nodes`: List of nodes, where each node is represented as `[node_id, resource_type]`
- `edges`: List of edges, where each edge is represented as `[source_node, target_node, edge_name]`

## Output Format
The output should provide the execution schedule of the program, indicating the start cycle of each operation. For example, the following output means that `n1` and `n2` start at cycle 0, while `n3` starts at cycle 3:
```
n1:0
n2:0
n3:3
```
\end{minted}
\section{Models}
\label{appendix:models}
The LLMs used in our experiments are listed in Table~\ref{tab:ablation}.
All models were accessed via official APIs provided by their respective organizations, except for the Meta models, which are accessed through the OpenRouter~\citep{openrouter} API.

\begin{table}[h]
\centering
\caption{Model specifications with API names and official pricing.}
\label{}
\resizebox{\linewidth}{!}{
\begin{tabular}{lllll}
\hline
\textbf{Organization} & \textbf{Model} & \textbf{API Name} & \textbf{Price (\$In/\$Out)} & \textbf{Type} \\
\hline
OpenAI & \GPTmini & o4-mini:high & 1.1/4.4 & Reasoning \\
Anthropic & \Claude & claude-3-7-sonnet-20250219 & 3/15 & Reasoning \\
DeepSeek & \DeepSeekV & deepseek-chat(0324) & 0.27/1.10 & Base \\
DeepSeek & \DeepSeekR & deepseek-reasoner & 0.55/2.19 & Reasoning \\
Google & \GeminiFlash & gemini-2.5-flash-preview-04-17 & 0.15/3.5 & Reasoning \\
Google & \GeminiPro & gemini-2.5-pro-preview-05-06 & 1.25/10.0 & Reasoning \\
Meta & \Llamas & meta-llama/Llama-3.3-70B-Instruct & 0.07/0.33 & Base \\
Meta & \Llamal & meta-llama/Llama-4-Maverick-17B-128E-Instruct & 0.27/0.85 & Base \\
Alibaba & \Qwen & qwen3-235b-a22b & 0.29/2.86 & Reasoning \\
\hline
\end{tabular}
}
\end{table}

\section{Problem Set}
\label{appendix:problem_set}
In this section, we provide more details on the problems included in Table~\ref{tab:problems}.
For a representative problem description used in the prompts, please consult our repository for additional details.

\subsection{Operator Scheduling}
Operator scheduling is a critical stage in high-level synthesis (HLS)~\citep{cong2011hls,pal2022dac}, the process of converting behavioral hardware descriptions into register-transfer level (RTL) implementations. This task involves carefully assigning each operation to a specific clock cycle while managing a variety of constraints such as data dependencies, resource availability, and performance targets. The effectiveness of the scheduling process is vital, as it directly influences key design metrics including area, power consumption, and execution time, making it an important focus in the field of electronic design automation (EDA).

Over the years, researchers have developed a wide range of techniques to tackle the inherent challenges of operator scheduling in HLS. Exact methods, such as those based on integer linear programming (ILP)~\citep{hwang1991ilphls,oppermann2016ilpmodulo}, can provide optimal solutions but often suffer from scalability issues. As a result, many commercial and academic HLS tools~\citep{xilinx_hls,canis2011legup} rely on heuristics to achieve practical, near-optimal results. Traditional heuristic approaches, including priority-function-based methods~\citep{shen2019entropyds,parker1986maha,paulin2002fds}, focus on balancing resource utilization with performance requirements. Notably, methods leveraging systems of difference constraints (SDC) enable an efficient formulation that captures a rich set of scheduling restrictions and casts the optimization objective into a linear programming (LP) framework~\citep{cong2006sdc,dai2018satsdc}. More recently, the incorporation of machine learning techniques~\citep{chen2019deeprl,liu2024diffopsched} has further advanced the state-of-the-art, enhancing both scheduling efficiency and solution quality in the face of increasingly complex hardware designs.

\subsection{Technology Mapping}
Technology mapping, in the context of logic synthesis for integrated circuits and field-programmable gate arrays (FPGAs), is the process of converting a logic network into an equivalent network of standard cells or logic resources from a specific technology library. The objective is to optimize key design metrics such as area, delay, and power consumption. It is a crucial step in the VLSI design flow and FPGA design flow, determining the actual physical implementation of a design. 

Here in our problem setting, we focus on area-optimal technology mapping for lookup table (LUT)-based FPGAs. 
Given an input logic network, the goal is to cover the network with $K$-input subgraphs, each of which can be implemented by a $K$-LUT, while minimizing the number of LUTs representing the circuit area. 

The most widely adopted approaches are cut-based methods, which operate in two stages: cut enumeration and cut selection. In this approach, all feasible $K$-input cuts---i.e., subgraphs with at most $K$ inputs---are enumerated for each node in the boolean network. Then, a dynamic programming-based selection process chooses one cut per node to construct a full LUT cover of the circuit, optimizing for metrics such as area or delay~\citep{chen2004daomap, cong1994flowmap, alan2006fpgatechmap}. 
A refinement of this approach is known as priority cut pruning, which retains only a limited set of the most promising cuts per node rather than considering all possible cuts. This significantly improves scalability for large circuits and is widely implemented in tools such as ABC~\citep{abc_tool}.

\subsection{Global Routing}
The global routing problem addresses the challenge of planning signal paths across a chip after logic placement, determining how a set of nets should traverse the layout to ensure connectivity while reserving space for detailed routing. Rather than producing exact wire geometries, global routing generates abstract paths through routing regions. This step must account for routing congestion, layer limitations, and timing criticality, while managing a growing number of nets in modern designs like Very-Large-Scale Integration (VLSI). The quality of the global routing solution plays a critical role in determining the feasibility and effectiveness of downstream routing stages and can ultimately dictate the success or failure of physical design closure.

The problem has been studied extensively via sequential and ILP-based methods. Maze routing, introduced by~\cite{lee2009algorithm}, laid the groundwork for sequential approaches, with subsequent improvements such as the work by~\cite{soukup1978fast}. For multi-terminal nets, rectilinear Steiner tree methods were developed~\citep{cong1998steiner}. However, sequential routing lacks global coordination and often leads to congestion. ILP-based methods formulate routing as a 0-1 programming, concurrently optimizing over all nets with objectives like wire length and capacity constraints. While exact ILP solvers are computationally intensive, relaxation techniques such as randomized rounding~\citep{carden1996global} and multi-commodity network flow models~\citep{shragowitz1987multicommodity, albrecht2001global} have been employed. Interior-point methods for solving the LP relaxation~\citep{vannelli2002adaptation, behjat2006integer} have also proven effective for scalable and near-optimal routing.

\cite{hu2001survey} provided a comprehensive survey of global routing techniques for integrated circuits. \cite{moffitt2009global} revisited the problem, offering a historical perspective and highlighting key open challenges that remain unresolved. More recently, to foster the development of advanced global routing methods, \cite{liang2024ispdcontest,liang2025ispd} introduced an ISPD contest that encourages the use of GPU-based techniques to accelerate global routing.

\subsection{E-Graph Extraction}
E-graph~\citep{chao1978chromatic, nelson1979simplification_toplas} is a data structure that compactly represents a set of expressions.
Given an input program and a set of rewrite rules, an e-graph is constructed by applying the rules to the program, generating new expressions, and merging equivalent expressions.
It has been widely used to represent and explore the huge number of equivalent program space in tensor graph transformation~\citep{tensat_mlsys,chen2024allo}, sparse linear algebra optimization~\citep{spores_vldb}, code optimization~\citep{laird2024speq_pldi, smith2024there}, digital signal processor (DSP) compilation~\citep{diospyros_asplos, thomas2024dsp_asplos}, circuit datapath synthesis~\citep{ustun2022impress_fccm, cheng2024seer_asplos}, and floating-point arithmetic~\citep{herbie_sigplan}.  

In an e-graph, all functionally equivalent terms are organized in the same equivalent classes, known as e-classes.
Nodes within each e-class that represent values or operators are called e-nodes.
E-classes are a partition of e-nodes, where each e-node belongs to exactly one e-class.
Dependencies in e-graphs are directed, which point from e-nodes to their children e-classes, indicating the operator (e-node) requires the values (e-nodes) from the child e-classes to compute its value.

In e-graph extraction, an optimized term from an e-graph is extracted after rewrites, based on a user-defined cost model.
The goal is to produce a functionally equivalent but improved implementation of the original input program.
The e-graph extraction problem is proven to be NP-hard when common sub-expressions are considered~\citep{eqsat_np, minsetcover}.

Existing e-graph extraction methods include exact methods employing ILP~\citep{cheng2024seer_asplos, smith2024there}.
Recently, there has been significant progress in employing heuristics for e-graph extraction.
These include a simple working-list method~\citep{herbie_sigplan}, a relaxation method utilizing gradient descent~\citep{cai2025egraph}, and a specialized method tailored for sparse e-graphs~\citep{goharshady2024fast}.
The dataset used in evaluation for this work primarily comes from SmoothE~\citep{cai2025egraph}.
\subsection{Intra-Operator Parallelism}
Intra-Operator Parallelism (IOPDDL), an emerging challenge introduced in the ASPLOS'25 contest track~\citep{moffit2025iopddl}, addresses the complexities of distributed deep learning.
Leading teams in this competition have predominantly employed metaheuristic approaches, distinguished by their unique pre-processing and optimization strategies.

The effective distribution of large machine learning models across multiple hardware accelerators is paramount for achieving desired performance in both training and serving applications~\citep{zheng2022alpa, zhao2023pytorch, shi2023tap, rajbhandari2020zero, lepikhin2020gshard, du2024liger,chen2024slapo}.
This task necessitates sharding the computation graph to minimize communication overhead, a process made intricate by the vast number of operations and tensors involved.
Specifically, for a given graph where nodes represent operations with distinct execution strategies (each possessing associated cost and memory usage), an optimal strategy must be chosen for every node.
The objective is to minimize the aggregate sum of node and edge costs, without exceeding a strict memory usage constraint across all devices at any point.
The inherent diversity in topological and memory characteristics of ML models across varied tasks and modalities renders this problem especially demanding.

\subsection{Protein Sequence Design}
Understanding how proteins fold into their native three-dimensional structures~\citep{jumper2021highly,watson2023novo} is a central problem in structural biology~\citep{min2022static,du2024machine}, traditionally framed as a forward problem: predicting the structure a given amino acid sequence will adopt~\citep{lin2023evolutionary,wang2024long}. In contrast, the protein sequence design or inverse folding problem starts from a fixed target structure and seeks sequences that are likely to fold into it. Many works have shown that this inverse formulation not only offers practical applications in protein engineering but also deepens our understanding of sequence–structure relationships~\citep{drexler1981molecular,yue1995forces,shakhnovich1993new,deutsch1996new,sun1995amino,lau1990theory}.

A common modeling approach treats sequence design as a global optimization problem over the space of amino acid sequences. Methods developed by~\cite{sun1995amino}, \cite{shakhnovich1993new}, and others define a fitness function to select sequences with favorable folding properties. These functions are designed to balance positive design (low free energy in the target structure) with negative design (high energy in competing folds), promoting both thermodynamic stability and structural specificity. More recently, people have been working on multi-state design with more or less general fitness functions~\citep{pokala2005energy,ambroggio2006computational,allen2010efficient,negron2013multistate,yanover2007dead,hallen2016comets,vucinic2020positive}.

In our benchmark, we focus on the Grand Canonical (GC) model~\citep{sun1995amino} of protein sequence design. The GC model operates on (i) a detailed three-dimensional geometric representation of a target structure with $n$ residues, (ii) a simplified binary alphabet distinguishing only hydrophobic (H) and polar (P) residues, and (iii) a fitness function $\Phi$ that favors sequences with densely packed hydrophobic cores while penalizing solvent-exposed hydrophobic residues. Despite its simplicity, the H/P model has been shown to capture key qualitative features of real protein structures~\citep{dill1995principles,kamtekar1993protein}. Several studies~\citep{micheletti1999protein,banavar1998structure} have explored the correspondence between sequences optimized under the GC model and those observed in natural proteins. However, a key obstacle has remained: computing an optimal sequence for a given structure is computationally challenging. The brute-force enumeration over all $2^n$ H/P sequences is infeasible for realistic protein sizes, and the algorithmic complexity of the problem was explicitly raised as an open question by~\cite{hart1997computational}. An efficient algorithm that constructs an optimal sequence in polynomial runtime was introduced later~\citep{kleinberg1999protein} using network flow.
\subsection{Mendelian Error Detection}
Chromosomes encode an individual's genetic information, with each gene occupying a specific position known as a locus. At each locus, a diploid organism carries two alleles---one inherited from each parent---forming its genotype. When direct genotyping is not available, researchers rely on the observable traits or phenotypes, which represent sets of compatible genotypes. A group of related individuals, along with their phenotypes at a locus, is organized into a pedigree, where each individual is either a founder or has parents defined within the structure.

Due to experimental and human errors, pedigree data may contain inaccuracies. These errors are classified as either parental errors (incorrect parentage, which we assume do not occur here) or phenotype errors, which can lead to Mendelian errors. A Mendelian error arises when all genotype combinations compatible with observed phenotypes violate Mendel's law that each individual inherits one allele from each parent. Detecting such inconsistencies is computationally challenging; the number of possible genotype combinations grows exponentially with pedigree size, making full enumeration impractical. In fact, verifying consistency has been shown to be NP-complete~\citep{aceto2004complexity}.

Error detection and correction are crucial for downstream tasks like genetic mapping or disease gene localization. However, existing tools are often limited by scalability issues, strong assumptions, or incomplete analysis. To address these limitations, a soft constraint network framework for detecting Mendelian inconsistencies was proposed~\citep{sanchez2008mendelian}, estimating the minimum number of required corrections, and suggesting optimal modifications. These problems naturally align with weighted constraint satisfaction and provide a rich testbed for scalable and flexible inference in large, complex pedigrees.
\subsection{Airline Crew Pairing}
The airline crew pairing problem is a well-established topic in operations research.
It involves constructing sequences of flight legs---known as pairings---that begin and end at a crew base, cover all scheduled flights, and satisfy a variety of regulatory and contractual constraints.
The primary goal is to minimize total crew-related costs, such as wages, hotel accommodations, and deadhead travel, while ensuring legality and operational feasibility.
This problem is typically formulated as a set partitioning model and addressed using column generation and branch-and-price techniques~\citep{desaulniers2006column, kasirzadeh2017airline}. Foundational systems developed for carriers like American Airlines demonstrated the effectiveness of these methods at scale~\citep{anbil1992global}.
More recent innovations include dynamic constraint aggregation~\citep{elhallaoui2005dynamicaggre} and machine learning-based pairing generation~\citep{yaakoubi2020mlcrew}, which are now integral to commercial solvers such as Jeppesen~\citep{jeppesen2021crew} and Sabre~\citep{sabre2020crew}, capable of processing monthly schedules with tens of thousands of flights.

In addition to exact methods, heuristic and metaheuristic techniques -- such as genetic algorithms, simulated annealing, and local search -- have been explored to improve scalability and reduce computation time, particularly for medium-sized instances or disruption recovery~\citep{luvcic2007metaheuristicscpp, souai2009geneticcpp}. These hybrid approaches aim to complement exact optimization methods by leveraging historical data and incorporating planner preferences, offering more flexible and adaptive solutions in practice.

\subsection{Pickup and Delivery Problem with Time Windows}

The Pick-up and Delivery Problem with Time Windows (PDPTW), originally proposed by~\cite{dumas1991pickup}, is generalized from a classical NP-hard combinatorial optimization problem---the Capacitated Vehicle Routing Problem (CVRP). It introduces additional complexity through precedence constraints, requiring pick-up locations to precede corresponding drop-off locations, and service time windows at each location. The problem can be seen in many logistic and public transportation systems, with the primary objective of minimizing the total travel cost. 

Over the past three decades, a wide range of models and algorithms have been proposed to address the PDPTW, with most falling into the category of heuristic or metaheuristic approaches. Prominent works include simulated annealing~\citep{li2001metapdptw, bent2006two}, large neighborhood search~\citep{curtois2018large, ropke2006adaptive}, and iterated local search~\citep{sartori2020study}. In contrast, research into exact solution methods has been relatively limited, with the most effective approaches relying on the set partitioning formulation combined with the branch-cut-and-price algorithm~\citep{ropke2009branch,baldacci2011exact}.
\cite{ropke2007models} provided a comprehensive survey of PDPTW solvers developed up to 2007.
~\cite{ho2018survey} later reviewed more recent advancements up to 2018, with a particular emphasis on PDPTW variants for people transportation, referred to as the Dial-a-Ride problem.
\cite{taniguchi2025timewindow} develop a mathematical model and applies heuristics (Genetic Algorithm, Simulated Annealing, Tabu Search) to analyze how time-window constraints affect urban pickup/delivery truck routing and scheduling.

To support algorithm development, several benchmark datasets have been created and maintained. The Li and Lim dataset~\citep{li2001metapdptw} is widely used and includes instances ranging from 100 to 1000 locations.
More recently, \cite{sartori2020study} released a larger-scale dataset generated from real-world spatial-temporal distributions.

\section{Additional Experiments}
\label{appendix:exp}
In this section, we provide more experimental results and analysis on our benchmark.

\subsection{Experimental Settings}
By default, we constrain LLMs to generate Python code for each problem and execute the code on a CPU server, with each instance allocated 8 CPU cores. The timeout for each problem is specified in Table~\ref{tab:timeout}.

\begin{table}[!htbp]
\caption{Timeout for each problem.}
\label{tab:timeout}
\centering
\small
\begin{tabular}{ll}\hline
\textbf{Problem} & \textbf{Timeout (sec)}\\\hline
Operator scheduling & 10\\
Technology mapping & 10\\
Global routing & 300\\\hline
E-graph extraction & 10\\
Intra-op parallelism & 60\\\hline
Protein sequence design & 10\\
Mendelian error detection & 10\\\hline
Airline crew pairing & 10\\
Pickup and delivery w/ time windows & 60\\\hline
\end{tabular}
\end{table}

\subsection{Detailed Results on Each Problem}
We provide the detailed \textsc{solve$_s$@$i$} values for each problem in Tables~\ref{tab:solvei_op} through~\ref{tab:solvei_pdptw}.
The variation in \textsc{solve$_s$@$i$} across different problems highlights the diverse levels of difficulty, as summarized in Table~\ref{tab:problems}.
For instance, the global routing problem remains unsolved by all evaluated LLMs -- even for generating a single feasible solution.
In the case of the pickup and delivery problem, the low \solve{3}{10} ratio also indicates that current LLMs struggle to consistently satisfy the problem's constraints.

\begin{table}[!htbp]
\caption{\textsc{solve$_s$@$i$} results on operator scheduling problem.}
\label{tab:solvei_op}
\centering
\resizebox{\linewidth}{!}{
\begin{tabular}{l|>{\columncolor{gray!20}}ccc|ccc|ccc}\hline
 &\multicolumn{3}{c|}{\textsc{solve$_\text{\RNum{3}}$}}
 &\multicolumn{3}{c|}{\textsc{solve$_\text{\RNum{2}}$}}
 &\multicolumn{3}{c}{\textsc{solve$_\text{\RNum{1}}$}}\\
\textbf{Model} & @10 & @5 & @1 & @10 & @5 & @1 & @10 & @5 & @1\\\hline
\DeepSeekV & \textbf{100.0\%} & \textbf{100.0\%} & 4.2\% & \textbf{100.0\%} & \textbf{100.0\%} & \textbf{100.0\%} & \textbf{100.0\%} & \textbf{100.0\%} & \textbf{100.0\%} \\
\DeepSeekR & \textbf{100.0\%} & \textbf{100.0\%} & \textbf{100.0\%} & \textbf{100.0\%} & \textbf{100.0\%} & \textbf{100.0\%} & \textbf{100.0\%} & \textbf{100.0\%} & \textbf{100.0\%} \\
\GeminiFlash & \textbf{100.0\%} & \textbf{100.0\%} & 0.0\% & \textbf{100.0\%} & \textbf{100.0\%} & 0.0\% & \textbf{100.0\%} & \textbf{100.0\%} & 0.0\% \\
\GeminiPro & \textbf{100.0\%} & \textbf{100.0\%} & \textbf{100.0\%} & \textbf{100.0\%} & \textbf{100.0\%} & \textbf{100.0\%} & \textbf{100.0\%} & \textbf{100.0\%} & \textbf{100.0\%} \\
\Llamal & 20.8\% & 0.0\% & 0.0\% & \textbf{100.0\%} & 4.2\% & 0.0\% & \textbf{100.0\%} & \textbf{100.0\%} & 4.2\% \\
\Llamas & \textbf{100.0\%} & \textbf{100.0\%} & \textbf{100.0\%} & \textbf{100.0\%} & \textbf{100.0\%} & \textbf{100.0\%} & \textbf{100.0\%} & \textbf{100.0\%} & \textbf{100.0\%} \\
\Qwen & \textbf{100.0\%} & \textbf{100.0\%} & \textbf{100.0\%} & \textbf{100.0\%} & \textbf{100.0\%} & \textbf{100.0\%} & \textbf{100.0\%} & \textbf{100.0\%} & \textbf{100.0\%} \\
\Claude & \textbf{100.0\%} & \textbf{100.0\%} & 0.0\% & \textbf{100.0\%} & \textbf{100.0\%} & 0.0\% & \textbf{100.0\%} & \textbf{100.0\%} & 0.0\% \\
\GPTmini & \textbf{100.0\%} & \textbf{100.0\%} & \textbf{100.0\%} & \textbf{100.0\%} & \textbf{100.0\%} & \textbf{100.0\%} & \textbf{100.0\%} & \textbf{100.0\%} & \textbf{100.0\%} \\\hline
\end{tabular}
}
\end{table}

\begin{table}[!htbp]
\caption{\textsc{solve$_s$@$i$} results on technology mapping problem.}
\label{tab:solvei_tech}
\centering
\resizebox{\linewidth}{!}{
\begin{tabular}{l|>{\columncolor{gray!20}}ccc|ccc|ccc}\hline
 &\multicolumn{3}{c|}{\textsc{solve$_\text{\RNum{3}}$}}
 &\multicolumn{3}{c|}{\textsc{solve$_\text{\RNum{2}}$}}
 &\multicolumn{3}{c}{\textsc{solve$_\text{\RNum{1}}$}}\\
\textbf{Model} & @10 & @5 & @1 & @10 & @5 & @1 & @10 & @5 & @1\\\hline
\DeepSeekV & 0.0\% & 0.0\% & 0.0\% & \textbf{100.0\%} & \textbf{100.0\%} & \textbf{100.0\%} & \textbf{100.0\%} & \textbf{100.0\%} & \textbf{100.0\%} \\
\DeepSeekR & 87.1\% & 87.1\% & \textbf{77.4\%} & \textbf{100.0\%} & \textbf{100.0\%} & \textbf{100.0\%} & \textbf{100.0\%} & \textbf{100.0\%} & \textbf{100.0\%} \\
\GeminiFlash & 0.0\% & 0.0\% & 0.0\% & 93.5\% & 77.4\% & 67.7\% & \textbf{100.0\%} & \textbf{100.0\%} & \textbf{100.0\%} \\
\GeminiPro & 74.2\% & 74.2\% & 0.0\% & \textbf{100.0\%} & \textbf{100.0\%} & 0.0\% & \textbf{100.0\%} & \textbf{100.0\%} & 0.0\% \\
\Llamal & 0.0\% & 0.0\% & 0.0\% & \textbf{100.0\%} & \textbf{100.0\%} & 0.0\% & \textbf{100.0\%} & \textbf{100.0\%} & 0.0\% \\
\Llamas & 0.0\% & 0.0\% & 0.0\% & \textbf{100.0\%} & \textbf{100.0\%} & 0.0\% & \textbf{100.0\%} & \textbf{100.0\%} & 6.5\% \\
\Qwen & 0.0\% & 0.0\% & 0.0\% & \textbf{100.0\%} & 87.1\% & 0.0\% & \textbf{100.0\%} & \textbf{100.0\%} & 3.2\% \\
\Claude & 87.1\% & 87.1\% & 0.0\% & \textbf{100.0\%} & \textbf{100.0\%} & 64.5\% & \textbf{100.0\%} & \textbf{100.0\%} & \textbf{100.0\%} \\
\GPTmini & \textbf{100.0\%} & \textbf{100.0\%} & 45.2\% & \textbf{100.0\%} & \textbf{100.0\%} & 51.6\% & \textbf{100.0\%} & \textbf{100.0\%} & \textbf{100.0\%} \\
\hline
\end{tabular}
}
\end{table}

\begin{table}[!htbp]
\caption{\textsc{solve$_s$@$i$} results on global routing problem.}
\label{tab:solvei_routing}
\centering
\resizebox{\linewidth}{!}{
\begin{tabular}{l|>{\columncolor{gray!20}}ccc|ccc|ccc}\hline
 &\multicolumn{3}{c|}{\textsc{solve$_\text{\RNum{3}}$}}
 &\multicolumn{3}{c|}{\textsc{solve$_\text{\RNum{2}}$}}
 &\multicolumn{3}{c}{\textsc{solve$_\text{\RNum{1}}$}}\\
\textbf{Model} & @10 & @5 & @1 & @10 & @5 & @1 & @10 & @5 & @1\\\hline
\DeepSeekV & \textbf{0.0\%} & \textbf{0.0\%} & \textbf{0.0\%} & 33.3\% & 33.3\% & 0.0\% & \textbf{100.0\%} & \textbf{100.0\%} & \textbf{100.0\%} \\
\DeepSeekR & \textbf{0.0\%} & \textbf{0.0\%} & \textbf{0.0\%} & 0.0\% & 0.0\% & 0.0\% & \textbf{100.0\%} & \textbf{100.0\%} & \textbf{100.0\%} \\
\GeminiFlash & \textbf{0.0\%} & \textbf{0.0\%} & \textbf{0.0\%} & 20.8\% & 0.0\% & 0.0\% & \textbf{100.0\%} & \textbf{100.0\%} & \textbf{100.0\%} \\
\GeminiPro & \textbf{0.0\%} & \textbf{0.0\%} & \textbf{0.0\%} & \textbf{100.0\%} & \textbf{100.0\%} & 0.0\% & \textbf{100.0\%} & \textbf{100.0\%} & 0.0\% \\
\Llamal & \textbf{0.0\%} & \textbf{0.0\%} & \textbf{0.0\%} & 0.0\% & 0.0\% & 0.0\% & 0.0\% & 0.0\% & 0.0\% \\
\Llamas & \textbf{0.0\%} & \textbf{0.0\%} & \textbf{0.0\%} & 0.0\% & 0.0\% & 0.0\% & \textbf{100.0\%} & \textbf{100.0\%} & 4.2\% \\
\Qwen & \textbf{0.0\%} & \textbf{0.0\%} & \textbf{0.0\%} & 0.0\% & 0.0\% & 0.0\% & \textbf{100.0\%} & \textbf{100.0\%} & \textbf{100.0\%} \\
\Claude & \textbf{0.0\%} & \textbf{0.0\%} & \textbf{0.0\%} & \textbf{100.0\%} & \textbf{100.0\%} & 0.0\% & \textbf{100.0\%} & \textbf{100.0\%} & 0.0\% \\
\GPTmini & \textbf{0.0\%} & \textbf{0.0\%} & \textbf{0.0\%} & \textbf{100.0\%} & \textbf{100.0\%} & \textbf{100.0\%} & \textbf{100.0\%} & \textbf{100.0\%} & \textbf{100.0\%} \\
\hline
\end{tabular}
}
\end{table}

\begin{table}[!htbp]
\caption{\textsc{solve$_s$@$i$} results on e-graph extraction problem.}
\label{tab:solvei_egraph}
\centering
\resizebox{\linewidth}{!}{
\begin{tabular}{l|>{\columncolor{gray!20}}ccc|ccc|ccc}\hline
 &\multicolumn{3}{c|}{\textsc{solve$_\text{\RNum{3}}$}}
 &\multicolumn{3}{c|}{\textsc{solve$_\text{\RNum{2}}$}}
 &\multicolumn{3}{c}{\textsc{solve$_\text{\RNum{1}}$}}\\
\textbf{Model} & @10 & @5 & @1 & @10 & @5 & @1 & @10 & @5 & @1\\\hline
\DeepSeekV & 4.3\% & 0.0\% & 0.0\% & \textbf{100.0\%} & \textbf{100.0\%} & 82.6\% & \textbf{100.0\%} & \textbf{100.0\%} & \textbf{100.0\%} \\
\DeepSeekR & \textbf{100.0\%} & \textbf{100.0\%} & \textbf{100.0\%} & \textbf{100.0\%} & \textbf{100.0\%} & \textbf{100.0\%} & \textbf{100.0\%} & \textbf{100.0\%} & \textbf{100.0\%} \\
\GeminiFlash & \textbf{100.0\%} & \textbf{100.0\%} & 0.0\% & \textbf{100.0\%} & \textbf{100.0\%} & 0.0\% & \textbf{100.0\%} & \textbf{100.0\%} & 0.0\% \\
\GeminiPro & \textbf{100.0\%} & \textbf{100.0\%} & 0.0\% & \textbf{100.0\%} & \textbf{100.0\%} & \textbf{100.0\%} & \textbf{100.0\%} & \textbf{100.0\%} & \textbf{100.0\%} \\
\Llamal & 0.0\% & 0.0\% & 0.0\% & \textbf{100.0\%} & \textbf{100.0\%} & 0.0\% & \textbf{100.0\%} & \textbf{100.0\%} & 0.0\% \\
\Llamas & 39.1\% & 39.1\% & 0.0\% & \textbf{100.0\%} & \textbf{100.0\%} & \textbf{100.0\%} & \textbf{100.0\%} & \textbf{100.0\%} & \textbf{100.0\%} \\
\Qwen & 87.0\% & 87.0\% & 87.0\% & \textbf{100.0\%} & \textbf{100.0\%} & \textbf{100.0\%} & \textbf{100.0\%} & \textbf{100.0\%} & \textbf{100.0\%} \\
\Claude & 39.1\% & 39.1\% & 0.0\% & \textbf{100.0\%} & \textbf{100.0\%} & \textbf{100.0\%} & \textbf{100.0\%} & \textbf{100.0\%} & \textbf{100.0\%} \\
\GPTmini & \textbf{100.0\%} & \textbf{100.0\%} & 39.1\% & \textbf{100.0\%} & \textbf{100.0\%} & \textbf{100.0\%} & \textbf{100.0\%} & \textbf{100.0\%} & \textbf{100.0\%} \\
\hline
\end{tabular}
}
\end{table}

\begin{table}[!htbp]
\caption{\textsc{solve$_s$@$i$} results on intra-op parallelism problem.}
\label{tab:solvei_iopddl}
\centering
\resizebox{\linewidth}{!}{
\begin{tabular}{l|>{\columncolor{gray!20}}ccc|ccc|ccc}\hline
 &\multicolumn{3}{c|}{\textsc{solve$_\text{\RNum{3}}$}}
 &\multicolumn{3}{c|}{\textsc{solve$_\text{\RNum{2}}$}}
 &\multicolumn{3}{c}{\textsc{solve$_\text{\RNum{1}}$}}\\
\textbf{Model} & @10 & @5 & @1 & @10 & @5 & @1 & @10 & @5 & @1\\\hline
\DeepSeekV & 82.1\% & 53.6\% & 35.7\% & 82.1\% & 53.6\% & 35.7\% & \textbf{100.0\%} & \textbf{100.0\%} & \textbf{100.0\%} \\
\DeepSeekR & 92.9\% & 92.9\% & 35.7\% & 92.9\% & 92.9\% & 35.7\% & \textbf{100.0\%} & \textbf{100.0\%} & 35.7\% \\
\GeminiFlash & \textbf{100.0\%} & \textbf{100.0\%} & \textbf{100.0\%} & \textbf{100.0\%} & \textbf{100.0\%} & \textbf{100.0\%} & \textbf{100.0\%} & \textbf{100.0\%} & \textbf{100.0\%} \\
\GeminiPro & 82.1\% & 82.1\% & 0.0\% & 82.1\% & 82.1\% & 0.0\% & \textbf{100.0\%} & \textbf{100.0\%} & 0.0\% \\
\Llamal & 96.4\% & 96.4\% & 3.6\% & \textbf{100.0\%} & \textbf{100.0\%} & 3.6\% & \textbf{100.0\%} & \textbf{100.0\%} & 3.6\% \\
\Llamas & 75.0\% & 75.0\% & 3.6\% & 82.1\% & 82.1\% & 3.6\% & \textbf{100.0\%} & \textbf{100.0\%} & \textbf{100.0\%} \\
\Qwen & 75.0\% & 71.4\% & 67.9\% & 78.6\% & 75.0\% & 75.0\% & \textbf{100.0\%} & \textbf{100.0\%} & \textbf{100.0\%} \\
\Claude & 82.1\% & 82.1\% & 71.4\% & 82.1\% & 82.1\% & 78.6\% & \textbf{100.0\%} & \textbf{100.0\%} & 96.4\% \\
\GPTmini & \textbf{100.0\%} & \textbf{100.0\%} & 92.9\% & \textbf{100.0\%} & \textbf{100.0\%} & \textbf{100.0\%} & \textbf{100.0\%} & \textbf{100.0\%} & \textbf{100.0\%} \\
\hline
\end{tabular}
}
\end{table}

\begin{table}[!htbp]
\caption{\textsc{solve$_s$@$i$} results on protein sequence design problem.}
\label{tab:solvei_protein}
\centering
\resizebox{\linewidth}{!}{
\begin{tabular}{l|>{\columncolor{gray!20}}ccc|ccc|ccc}\hline
 &\multicolumn{3}{c|}{\textsc{solve$_\text{\RNum{3}}$}}
 &\multicolumn{3}{c|}{\textsc{solve$_\text{\RNum{2}}$}}
 &\multicolumn{3}{c}{\textsc{solve$_\text{\RNum{1}}$}}\\
\textbf{Model} & @10 & @5 & @1 & @10 & @5 & @1 & @10 & @5 & @1\\\hline
\DeepSeekV & 83.3\% & 83.3\% & 83.3\% & \textbf{100.0\%} & \textbf{100.0\%} & \textbf{100.0\%} & \textbf{100.0\%} & \textbf{100.0\%} & \textbf{100.0\%} \\
\DeepSeekR & 87.5\% & 87.5\% & 0.0\% & \textbf{100.0\%} & \textbf{100.0\%} & 0.0\% & \textbf{100.0\%} & \textbf{100.0\%} & 0.0\% \\
\GeminiFlash & 95.8\% & \textbf{95.8\%} & \textbf{95.8\%} & \textbf{100.0\%} & \textbf{100.0\%} & \textbf{100.0\%} & \textbf{100.0\%} & \textbf{100.0\%} & \textbf{100.0\%} \\
\GeminiPro & \textbf{100.0\%} & \textbf{95.8\%} & 0.0\% & \textbf{100.0\%} & 95.8\% & 0.0\% & \textbf{100.0\%} & \textbf{100.0\%} & 4.2\% \\
\Llamal & 83.3\% & 83.3\% & 0.0\% & 95.8\% & 95.8\% & 0.0\% & \textbf{100.0\%} & \textbf{100.0\%} & 4.2\% \\
\Llamas & 12.5\% & 12.5\% & 12.5\% & 95.8\% & 95.8\% & 95.8\% & 95.8\% & 95.8\% & 95.8\% \\
\Qwen & 87.5\% & 87.5\% & 87.5\% & \textbf{100.0\%} & \textbf{100.0\%} & \textbf{100.0\%} & \textbf{100.0\%} & \textbf{100.0\%} & \textbf{100.0\%} \\
\Claude & 58.3\% & 45.8\% & 0.0\% & \textbf{100.0\%} & \textbf{100.0\%} & 0.0\% & \textbf{100.0\%} & \textbf{100.0\%} & 0.0\% \\
\GPTmini & 91.7\% & 91.7\% & 91.7\% & \textbf{100.0\%} & \textbf{100.0\%} & \textbf{100.0\%} & \textbf{100.0\%} & \textbf{100.0\%} & \textbf{100.0\%} \\
\hline
\end{tabular}
}
\end{table}

\begin{table}[!htbp]
\caption{\textsc{solve$_s$@$i$} results on mendelian error detection problem.}
\label{tab:solvei_pedigree}
\centering
\resizebox{\linewidth}{!}{
\begin{tabular}{l|>{\columncolor{gray!20}}ccc|ccc|ccc}\hline
 &\multicolumn{3}{c|}{\textsc{solve$_\text{\RNum{3}}$}}
 &\multicolumn{3}{c|}{\textsc{solve$_\text{\RNum{2}}$}}
 &\multicolumn{3}{c}{\textsc{solve$_\text{\RNum{1}}$}}\\
\textbf{Model} & @10 & @5 & @1 & @10 & @5 & @1 & @10 & @5 & @1\\\hline
\DeepSeekV & \textbf{100.0\%} & \textbf{100.0\%} & 0.0\% & \textbf{100.0\%} & \textbf{100.0\%} & 0.0\% & \textbf{100.0\%} & \textbf{100.0\%} & 0.0\% \\
\DeepSeekR & \textbf{100.0\%} & \textbf{100.0\%} & 0.0\% & \textbf{100.0\%} & \textbf{100.0\%} & 0.0\% & \textbf{100.0\%} & \textbf{100.0\%} & 0.0\% \\
\GeminiFlash & \textbf{100.0\%} & 10.0\% & 10.0\% & \textbf{100.0\%} & \textbf{100.0\%} & \textbf{100.0\%} & \textbf{100.0\%} & \textbf{100.0\%} & \textbf{100.0\%} \\
\GeminiPro & 80.0\% & 80.0\% & \textbf{80.0\%} & 80.0\% & 80.0\% & 80.0\% & \textbf{100.0\%} & \textbf{100.0\%} & \textbf{100.0\%} \\
\Llamal & 60.0\% & 60.0\% & 60.0\% & 60.0\% & 60.0\% & 60.0\% & 60.0\% & 60.0\% & 60.0\% \\
\Llamas & 55.0\% & 55.0\% & 55.0\% & 55.0\% & 55.0\% & 55.0\% & \textbf{100.0\%} & \textbf{100.0\%} & \textbf{100.0\%} \\
\Qwen & 55.0\% & 55.0\% & 0.0\% & \textbf{100.0\%} & \textbf{100.0\%} & 0.0\% & \textbf{100.0\%} & \textbf{100.0\%} & 0.0\% \\
\Claude & \textbf{100.0\%} & \textbf{100.0\%} & 0.0\% & \textbf{100.0\%} & \textbf{100.0\%} & \textbf{100.0\%} & \textbf{100.0\%} & \textbf{100.0\%} & \textbf{100.0\%} \\
\GPTmini & \textbf{100.0\%} & 50.0\% & 35.0\% & \textbf{100.0\%} & \textbf{100.0\%} & \textbf{100.0\%} & \textbf{100.0\%} & \textbf{100.0\%} & \textbf{100.0\%} \\
\hline
\end{tabular}
}
\end{table}

\begin{table}[!htbp]
\caption{\textsc{solve$_s$@$i$} results on airline crew pairing problem.}
\label{tab:solvei_cpp}
\centering
\resizebox{\linewidth}{!}{
\begin{tabular}{l|>{\columncolor{gray!20}}ccc|ccc|ccc}\hline
 &\multicolumn{3}{c|}{\textsc{solve$_\text{\RNum{3}}$}}
 &\multicolumn{3}{c|}{\textsc{solve$_\text{\RNum{2}}$}}
 &\multicolumn{3}{c}{\textsc{solve$_\text{\RNum{1}}$}}\\
\textbf{Model} & @10 & @5 & @1 & @10 & @5 & @1 & @10 & @5 & @1\\\hline
\DeepSeekV & \textbf{100.0\%} & \textbf{100.0\%} & 0.0\% & \textbf{100.0\%} & \textbf{100.0\%} & \textbf{100.0\%} & \textbf{100.0\%} & \textbf{100.0\%} & \textbf{100.0\%} \\
\DeepSeekR & \textbf{100.0\%} & \textbf{100.0\%} & \textbf{100.0\%} & \textbf{100.0\%} & \textbf{100.0\%} & \textbf{100.0\%} & \textbf{100.0\%} & \textbf{100.0\%} & \textbf{100.0\%} \\
\GeminiFlash & 0.0\% & 0.0\% & 0.0\% & 0.0\% & 0.0\% & 0.0\% & \textbf{100.0\%} & \textbf{100.0\%} & 14.3\% \\
\GeminiPro & 0.0\% & 0.0\% & 0.0\% & 0.0\% & 0.0\% & 0.0\% & \textbf{100.0\%} & \textbf{100.0\%} & \textbf{100.0\%} \\
\Llamal & \textbf{100.0\%} & \textbf{100.0\%} & 0.0\% & \textbf{100.0\%} & \textbf{100.0\%} & 35.7\% & \textbf{100.0\%} & \textbf{100.0\%} & \textbf{100.0\%} \\
\Llamas & 42.9\% & 42.9\% & 42.9\% & 42.9\% & 42.9\% & 42.9\% & \textbf{100.0\%} & \textbf{100.0\%} & \textbf{100.0\%} \\
\Qwen & 21.4\% & 21.4\% & 0.0\% & \textbf{100.0\%} & 85.7\% & 0.0\% & \textbf{100.0\%} & \textbf{100.0\%} & 0.0\% \\
\Claude & \textbf{100.0\%} & \textbf{100.0\%} & 0.0\% & \textbf{100.0\%} & \textbf{100.0\%} & 0.0\% & \textbf{100.0\%} & \textbf{100.0\%} & 0.0\% \\
\GPTmini & \textbf{100.0\%} & \textbf{100.0\%} & \textbf{100.0\%} & \textbf{100.0\%} & \textbf{100.0\%} & \textbf{100.0\%} & \textbf{100.0\%} & \textbf{100.0\%} & \textbf{100.0\%} \\
\hline
\end{tabular}
}
\end{table}

\begin{table}[!htbp]
\caption{\textsc{solve$_s$@$i$} results on pickup and delivery with time windows problem.}
\label{tab:solvei_pdptw}
\centering
\resizebox{\linewidth}{!}{
\begin{tabular}{l|>{\columncolor{gray!20}}ccc|ccc|ccc}\hline
 &\multicolumn{3}{c|}{\textsc{solve$_\text{\RNum{3}}$}}
 &\multicolumn{3}{c|}{\textsc{solve$_\text{\RNum{2}}$}}
 &\multicolumn{3}{c}{\textsc{solve$_\text{\RNum{1}}$}}\\
\textbf{Model} & @10 & @5 & @1 & @10 & @5 & @1 & @10 & @5 & @1\\\hline
\DeepSeekV & 0.0\% & 0.0\% & 0.0\% & 80.0\% & 73.3\% & 73.3\% & \textbf{100.0\%} & \textbf{100.0\%} & \textbf{100.0\%} \\
\DeepSeekR & 16.7\% & 13.3\% & 3.3\% & \textbf{100.0\%} & \textbf{100.0\%} & \textbf{100.0\%} & \textbf{100.0\%} & \textbf{100.0\%} & \textbf{100.0\%} \\
\GeminiFlash & \textbf{96.7\%} & \textbf{90.0\%} & 6.7\% & \textbf{100.0\%} & \textbf{100.0\%} & \textbf{100.0\%} & \textbf{100.0\%} & \textbf{100.0\%} & \textbf{100.0\%} \\
\GeminiPro & 30.0\% & 26.7\% & \textbf{13.3\%} & \textbf{100.0\%} & \textbf{100.0\%} & \textbf{100.0\%} & \textbf{100.0\%} & \textbf{100.0\%} & \textbf{100.0\%} \\
\Llamal & 0.0\% & 0.0\% & 0.0\% & \textbf{100.0\%} & \textbf{100.0\%} & 0.0\% & \textbf{100.0\%} & \textbf{100.0\%} & 0.0\% \\
\Llamas & 0.0\% & 0.0\% & 0.0\% & \textbf{100.0\%} & \textbf{100.0\%} & 0.0\% & \textbf{100.0\%} & \textbf{100.0\%} & 0.0\% \\
\Qwen & 0.0\% & 0.0\% & 0.0\% & \textbf{100.0\%} & \textbf{100.0\%} & \textbf{100.0\%} & \textbf{100.0\%} & \textbf{100.0\%} & \textbf{100.0\%} \\
\Claude & 0.0\% & 0.0\% & 0.0\% & \textbf{100.0\%} & \textbf{100.0\%} & 16.7\% & \textbf{100.0\%} & \textbf{100.0\%} & \textbf{100.0\%} \\
\GPTmini & 3.3\% & 0.0\% & 0.0\% & \textbf{100.0\%} & \textbf{100.0\%} & \textbf{100.0\%} & \textbf{100.0\%} & \textbf{100.0\%} & \textbf{100.0\%} \\
\hline
\end{tabular}
}
\end{table}

\newpage
\subsection{Ablation on Temperature}
\label{appendix:temperature}
We evaluate various models across different temperature settings, $T \in \{0.0, 0.5, 1.0\}$. For each model, we run 10 iterations per problem and report the highest QYI achieved across these iterations as the final QYI score for that problem. The overall benchmark score is then computed as the arithmetic mean of QYI across all problems. Detailed results are shown in Tables~\ref{tab:t0} to~\ref{tab:t1}.

In general, improving the temperature can be beneficial to quality as the model becomes more creative, but may harm yield as it may not follow the constraints strictly.
Note that yield emphasizes the best iteration that achieves the highest QYI, whereas \textsc{solve}$_{\text{\RNum{3}}}$ reflects the cumulative success rate across iterations; therefore, their values may differ.
Additionally, the weighted QYI is not the harmonic mean of weighted yield and weighted quality, as it is computed by aggregating metrics across different problems using a weighted approach.

We also report an uncapped version of the weighted QYI metric\footnote{The uncapped version of quality is computed as $1/\hat{N} \sum_{n=1}^{\hat{N}} c^\star_n / c_n$, and the uncapped QYI is derived by substituting the original quality metric with this uncapped variant.}, which better reflects cases where LLM-generated programs outperform expert solutions on certain test instances. Improvements are underlined in the tables. While this variant achieves slightly higher scores for most models -- indicating occasional superior performance -- it also confirms that, in the majority of cases, LLMs still lag significantly behind expert solutions.

\begin{table}[!htbp]
\caption{Performance of different models on $\text{Temperature}=0$.}
\label{tab:t0}
\centering
\resizebox{\linewidth}{!}{
\begin{tabular}{lcccc}\hline
\textbf{Model} &  \textbf{Weighted Yield} & \textbf{Weighted Quality} & \makecell[c]{\textbf{Weighted QYI}\\\textbf{(Capped)}} & \makecell[c]{\textbf{Weighted QYI}\\\textbf{(Uncapped)}}\\\hline
\Claude &  0.5963  &         0.4686    &      0.5034  &     0.5034  \\
\DeepSeekR       &  \textbf{0.6972}  &         0.5775    &      0.5498  &     \underline{0.5553}  \\
\DeepSeekV       &  0.4587  &         0.3890    &      0.3707 &     0.3707 \\
\GeminiFlash  &  0.6606  &         0.5281    &      0.5682  &     \underline{0.5753}  \\
\GeminiPro    &  0.6468  &         \textbf{0.6700}    &      \textbf{0.6170}  &     \textbf{\underline{0.6228}}  \\
\Llamas     &  0.3394  &         0.3521    &      0.2951  &     0.2953  \\
\Llamal  &  0.3211  &         0.3383    &      0.2955  &     0.2955  \\
\Qwen        &  0.4450  &         0.4513    &      0.4355  &     \underline{0.4423}\\\hline
\end{tabular}
}
\end{table}

\begin{table}[!htbp]
\caption{Performance of different models on $\text{Temperature}=0.5$.}
\label{tab:t05}
\centering
\resizebox{\linewidth}{!}{
\begin{tabular}{lcccc}\hline
\textbf{Model} &  \textbf{Weighted Yield} & \textbf{Weighted Quality} & \makecell[c]{\textbf{Weighted QYI}\\\textbf{(Capped)}} & \makecell[c]{\textbf{Weighted QYI}\\\textbf{(Uncapped)}}\\\hline
\Claude   &  \textbf{0.6147}   &        \textbf{0.6468}    &       \textbf{0.5437} & \underline{\textbf{0.5451}}\\       
\DeepSeekR         &  0.5138   &        0.5751    &       0.4743 & \underline{0.4812}\\       
\DeepSeekV         &  0.3716   &        0.4645    &       0.3322 & 0.3322\\            
\GeminiFlash    &  0.4817   &        0.5700    &       0.4760 & \underline{0.4828}\\       
\GeminiPro      &  0.4817   &        0.5609    &       0.4767 & \underline{0.4789}\\       
\Llamas       &  0.3991   &        0.4407    &       0.4108 & 0.4108\\       
\Llamal    &  0.3349   &        0.3712    &       0.3050 & \underline{0.3646}\\       
\Qwen          &  0.4128   &        0.4798    &       0.4269 & \underline{0.4327}\\\hline       
\end{tabular}
}
\end{table}

\begin{table}[!htbp]
\caption{Performance of different models on $\text{Temperature}=1$.}
\label{tab:t1}
\centering
\resizebox{\linewidth}{!}{
\begin{tabular}{lcccc}\hline
\textbf{Model} &  \textbf{Weighted Yield} & \textbf{Weighted Quality} & \makecell[c]{\textbf{Weighted QYI}\\\textbf{(Capped)}} & \makecell[c]{\textbf{Weighted QYI}\\\textbf{(Uncapped)}}\\\hline
\Claude   &  0.5138   &        0.5924    &       0.4828 & \underline{0.4841}\\       
\DeepSeekR         &  0.5688   &        0.5625    &       0.5313 & \underline{0.5383}\\       
\DeepSeekV         &  0.4128   &        0.4188    &       0.3839 & \underline{0.3841}\\       
\GPTmini         &  \textbf{0.6927}   &        0.6440    &       \textbf{0.6089} & \underline{\textbf{0.6158}}\\       
\GeminiFlash    &  0.4771   &        \textbf{0.7688}    &       0.5030 & \underline{0.5047}\\       
\GeminiPro      &  0.5229   &        0.4893    &       0.4921 & \underline{0.4981}\\       
\Llamas       &  0.3028   &        0.3627    &       0.2868 & \underline{0.2916}\\       
\Llamal    &  0.2982   &        0.3271    &       0.2667 & \underline{0.2672}\\       
\Qwen          &  0.5459   &        0.5228    &       0.5294 & \underline{0.5364}\\\hline  
\end{tabular}
}
\end{table}

\subsection{Few-Shot Demonstration}
\label{appendix:few-shot}
Table~\ref{tab:fewshot_appendix} highlights the impact of few-shot demonstrations on LLM performance across the entire \Name benchmark. Introducing only a small number of demonstrations (e.g., three) can negatively affect solution quality and success rate, as these examples may not be representative of the overall dataset, leading the model to overfit to them. However, providing a larger set of demonstrations can potentially improve QYI, as the model benefits from greater diversity and can learn more generalizable patterns.

\begin{table}[!htbp]
\caption{Impact of few-shot demonstrations on performance (Model: \GeminiPro).}
\label{tab:fewshot_appendix}
\centering
\small
\begin{tabular}{cccc}\hline
\textbf{\# of Demos} & \textbf{Weighted Yield} & \textbf{Weighted Quality} & \textbf{Weighted QYI}\\\hline
Zero-shot  & 0.5872     &      0.7159     &      0.5999\\
Half-shot  & 0.5092     &      0.6526     &      0.5361\\
Full-shot  & 0.6468     &      0.6700     &      0.6170\\\hline
\end{tabular}
\end{table}

\subsection{Feedback Rounds}
\label{appendix:feedback-rounds}
Table~\ref{tab:feedback_appendix} shows that increasing the number of feedback rounds has a nuanced impact on performance. While a moderate number of rounds (e.g., five) can enhance overall quality by guiding the model to refine its solutions, excessive feedback may lead to diminishing returns or even degrade performance. This suggests that too many rounds can overwhelm the model, making it harder to identify and prioritize the most critical information from the feedback.

\begin{table}[!htbp]
\caption{Impact of feedback rounds on performance (Model: \GeminiPro).}
\label{tab:feedback_appendix}
\centering
\small
\begin{tabular}{cccc}\hline
\textbf{\# of Feedback Rounds} & \textbf{Weighted Yield} & \textbf{Weighted Quality} & \textbf{Weighted QYI}\\\hline
1 & 0.6193     &      0.7290      &     0.6253\\
5 & 0.6055     &      0.7313      &     0.6259\\
10 & 0.6468    &       0.6700     &      0.6170\\\hline
\end{tabular}
\end{table}

\subsection{Iterative Best-of-N Sampling}
\label{appendix:bestofn}
To investigate the benefits of test-time search strategies, we sample $k$ candidate programs in each iteration, evaluate them, and return feedback for all $k$ programs to the LLM. After a fixed number of iterations, we select the best-performing program from the entire pool -- a process we refer to as \emph{iterative best-of-N sampling}. The total number of sampled programs is held constant across different values of $k$. This strategy allows the model to explore diverse candidate solutions in parallel and evolve the program based on evaluative feedback.

As shown in Table~\ref{tab:ksample_appendix}, increasing $k$ leads to better quality of results, indicating that aggregating feedback across multiple candidates allows the LLM to better explore the solution space and improve sampling efficiency by allocating computational budget toward more informative evaluations.

\begin{table}[!htbp]
\caption{Impact of iterative best-of-N sampling on performance (Model: \GeminiPro).}
\label{tab:ksample_appendix}
\centering
\small
\begin{tabular}{cccc}\hline
\textbf{\# of Samples @ Iteration} & \textbf{Weighted Yield} & \textbf{Weighted Quality} & \textbf{Weighted QYI}\\\hline
2@5 &  0.5688      &     0.7698       &    0.6160\\\hline
1@10  & 0.6468     &      0.6700      &     0.6170\\\hline
\end{tabular}
\end{table}

\subsection{Error Analysis}
\label{appendix:errors}
In the following, we present representative examples of common errors made by LLMs during heuristic generation. These errors highlight current limitations in code reliability and execution:
\begin{itemize}
\item \textbf{Import error:}
This type of error occurs when the generated code relies on external libraries that are not available in the environment. In the example below, the model attempts to import the \texttt{ortools} library, which results in a \texttt{ModuleNotFoundError}. Such errors suggest that the model does not strictly follow the instructions given in the prompt.
\begin{minted}[breaklines=true, breakanywhere=true,fontsize=\small, frame=single]{python}
  File "operator_scheduling/gemini-2.5-flash-preview-04-17/iteration4/solver.py", line 2, in <module>
    from ortools.sat.python import cp_model
ModuleNotFoundError: No module named 'ortools'
\end{minted}

\item \textbf{API misuse error:}
LLMs often misuse APIs due to a misunderstanding of library interfaces. In the following case, the model tries to call \texttt{random()} directly from the \texttt{random} module, which is not callable.
\begin{minted}[breaklines=true, breakanywhere=true,fontsize=\small, frame=single]{python}
  File "intra_op_parallel/o4-mini/iteration3/solver.py", line 64, in init_jitter
    if len(ci) > 1 and random() < 0.1:
                       ^^^^^^^^
TypeError: 'module' object is not callable
\end{minted}

\item \textbf{Syntax error:}
Syntax errors are common when the model fails to adhere to basic language rules. In this example, there is an unmatched parenthesis in a \texttt{while} loop condition, leading to a \texttt{SyntaxError}. Such mistakes typically indicate a lack of code completion validation in the generation process.
\begin{minted}[breaklines=true, breakanywhere=true,fontsize=\small, frame=single]{python}
  File "crew_pairing/deepseek-chat/iteration7/solver.py", line 60
    while len(used_legs) < len(df)):
                                  ^
SyntaxError: unmatched ')'
\end{minted}

\item \textbf{Runtime error:}
Even syntactically and semantically correct code can fail at runtime. In this case, the model modifies a dictionary while iterating over it, which raises a \texttt{RuntimeError}. This highlights the model's difficulty in reasoning about the actual executable code in a long context.
\begin{minted}[breaklines=true, breakanywhere=true,fontsize=\small, frame=single]{python}
  File "technology_mapping/llama-4-maverick/iteration2/solver.py", line 104, in technology_mapping
    for successor in G.successors(node):
RuntimeError: dictionary changed size during iteration
\end{minted}
\end{itemize}

\subsection{C++ Example}
\label{appendix:cpp}
We conduct preliminary experiments on the technology mapping problem by modifying the prompt to instruct the LLM to generate a C++ solution, using the provided function template:
\texttt{void solve(const std::string\& input\_file, const std::string\& output\_file)}. 

Integrating C++ into our agentic feedback loop remains challenging due to dependencies on domain-specific libraries and the complexity of parallel execution.
As a result, our preliminary experiment with C++ involves only a single iteration of prompting. 

Table~\ref{tab:c++_appendix} presents a performance comparison between the Python solution with 10 iterations and the C++ solution with just one iteration. 
Although the C++ solution does not produce high-quality output in its initial attempt, it already achieves a better yield than the Python solution after 10 iterations -- an unexpectedly strong outcome.
Notably, the Python solution fails to generate any valid result in its first iteration.
This is attributed to the significantly faster execution speed of C++ code, which enables it to avoid the timeout errors frequently encountered by Python in this task.

We expect to see further performance improvement with C++ after we integrate it into the feedback loop in our framework.

\begin{table}[!htbp]
\caption{Impact of C++ code on technology mapping performance (Model: \texttt{Gemini-2.5-pro}).}
\label{tab:c++_appendix}
\centering
\begin{tabular}{ccccc}\hline
\textbf{Language} & \textbf{\# of Iterations}  & \textbf{Yield} & \textbf{Quality} & \textbf{QYI}\\\hline
Python & 10 & 0.7419  &  0.6423 & 0.6885\\
C++     & 1  & 0.7742  & 0.3493  & 0.4814 \\\hline
\end{tabular}
\end{table}

\subsection{Token Usage}
Table~\ref{tab:token} presents an example of token usage when running the complete \Name benchmark across different models. Among them, \GeminiPro consumes the most tokens for prompt and completion.

\begin{table}[h]
\centering
\caption{Token counts from a single run of \Name across different models.}
\label{tab:token}
\begin{tabular}{lllll}
\hline
\textbf{Model} & \textbf{Prompt Tokens} & \textbf{Completion Tokens}\\\hline
\Claude & 1911947 & 298369 \\
\DeepSeekR & 1215661 & 702746  \\
\DeepSeekV & 1536084 & 122058  \\
\GPTmini & 1063688 & 978891  \\
\GeminiFlash & 2742487 & 332569  \\
\GeminiPro  & 2880737 & 455739  \\
\Llamas &  969887  & 77523  \\
\Llamal & 1497730 &  77080  \\ 
\Qwen &  1089888 &  127447 \\\hline
\end{tabular}
\end{table}

\section{Detailed Analysis of Case Study}
\label{appendix:case-study}

Here we include a detailed analysis of the case study in Section~\ref{subsec:case_study}. 
We append the entire code of each iteration and analyze the generated solution.

In the first iteration, the \GPTmini model generates an algorithm based on cut enumeration and dynamic programming (DP)-based cut selection, which is the common approach used in technology mapping tools. The core logic is highlighted in the following code snippet. 
However, it does not prune the cut enumeration space and computes all possible $K$-feasible cuts for each node, which leads to an exponential explosion in computation time. 
While this achieves high quality, it frequently triggers \texttt{TimeoutError} for larger datasets, resulting in low yield. 

\begin{minted}[breaklines=true, breakanywhere=true,fontsize=\scriptsize, linenos, frame=single, highlightcolor=yellow!35, highlightlines={153-175, 179-203}]{python}
# iteration 1: base algorithm without pruning the search space

def solve(input_file: str, solution_file: str):
    from collections import deque
    class Node:
        __slots__ = ('name','fanin_names','fanins','fanouts','patterns_one','patterns_zero','const_val')
        def __init__(self,name):
            self.name = name
            self.fanin_names = []
            self.fanins = []
            self.fanouts = []
            self.patterns_one = []
            self.patterns_zero = []
            self.const_val = None
    raw_lines = open(input_file).read().splitlines()
    lines = []
    i = 0
    while i < len(raw_lines):
        line = raw_lines[i].rstrip()
        if not line or line.lstrip().startswith('#'):
            i += 1
            continue
        if line.endswith('\\'):
            merged = line[:-1].rstrip()
            i += 1
            while i < len(raw_lines):
                nxt = raw_lines[i].rstrip()
                if nxt.endswith('\\'):
                    merged += ' ' + nxt[:-1].rstrip()
                    i += 1
                else:
                    merged += ' ' + nxt
                    i += 1
                    break
            lines.append(merged)
        else:
            lines.append(line)
            i += 1
    model_name = ''
    input_list = []
    output_list = []
    nodes = {}
    idx = 0
    while idx < len(lines):
        l = lines[idx].strip()
        if not l or l.startswith('#'):
            idx += 1
            continue
        if l.startswith('.model'):
            parts = l.split()
            if len(parts) > 1:
                model_name = parts[1]
            idx += 1
        elif l.startswith('.inputs'):
            parts = l.split()
            for name in parts[1:]:
                input_list.append(name)
                if name not in nodes:
                    nodes[name] = Node(name)
            idx += 1
        elif l.startswith('.outputs'):
            parts = l.split()
            for name in parts[1:]:
                output_list.append(name)
            idx += 1
        elif l.startswith('.names'):
            parts = l.split()
            in_names = parts[1:-1]
            out_name = parts[-1]
            if out_name not in nodes:
                nodes[out_name] = Node(out_name)
            node = nodes[out_name]
            node.fanin_names = in_names[:]
            patterns = []
            j = idx + 1
            while j < len(lines):
                lj = lines[j].strip()
                if not lj or lj.startswith('.') or lj.startswith('#'):
                    break
                patterns.append(lj)
                j += 1
            if len(in_names) == 0:
                if patterns and patterns[0].strip() == '1':
                    node.const_val = 1
                else:
                    node.const_val = 0
            else:
                for pline in patterns:
                    parts2 = pline.split()
                    if len(parts2) == 1:
                        mask = parts2[0]; bit = '1'
                    else:
                        mask = parts2[0]; bit = parts2[1]
                    if bit == '1':
                        node.patterns_one.append(mask)
                    else:
                        node.patterns_zero.append(mask)
            idx = j
        elif l.startswith('.end'):
            break
        else:
            idx += 1
    for node in nodes.values():
        node.fanins = []
        for fn in node.fanin_names:
            if fn in nodes:
                node.fanins.append(nodes[fn])
        node.fanouts = []
    for node in nodes.values():
        for f in node.fanins:
            f.fanouts.append(node)
    reachable = set()
    dq = deque()
    for out in output_list:
        if out in nodes:
            dq.append(out)
    while dq:
        u = dq.pop()
        if u in reachable:
            continue
        reachable.add(u)
        for f in nodes[u].fanins:
            if f.name not in reachable:
                dq.append(f.name)
    nodes = {name:node for name,node in nodes.items() if name in reachable}
    for node in nodes.values():
        node.fanins = [f for f in node.fanins if f.name in nodes]
        node.fanouts = [f for f in node.fanouts if f.name in nodes]
    indeg = {name: len(node.fanins) for name,node in nodes.items()}
    dq = deque([name for name,d in indeg.items() if d == 0])
    topo_names = []
    while dq:
        u = dq.popleft()
        topo_names.append(u)
        for w in nodes[u].fanouts:
            indeg[w.name] -= 1
            if indeg[w.name] == 0:
                dq.append(w.name)
    topo_list = [nodes[name] for name in topo_names]
    K = 6
    def prune_cuts(cset):
        cuts = list(cset)
        res = []
        for c in cuts:
            skip = False
            for d in cuts:
                if d is not c and d.issubset(c):
                    skip = True
                    break
            if not skip:
                res.append(c)
        return res
    cuts = {}      # Cut Enumeration
    for n in topo_list:
        if not n.fanins:
            cuts[n.name] = [frozenset([n.name])]
        else:
            cuts_n = None
            for f in n.fanins:
                cf = cuts[f.name]
                if cuts_n is None:
                    cuts_n = cf[:]
                else:
                    newset = set()
                    for c1 in cuts_n:
                        for c2 in cf:
                            u = c1 | c2
                            if len(u) <= K:
                                newset.add(u)
                    cuts_n = prune_cuts(newset)
            if cuts_n is None:
                cuts_n = []
            cuts_set = set(cuts_n)
            cuts_set.add(frozenset([n.name]))
            cuts[n.name] = prune_cuts(cuts_set)
    cost = {}
    best_cut = {}
    INF = 10**18
    for n in topo_list:    # DP-based Cut Selection
        if not n.fanins:
            cost[n.name] = 0
            best_cut[n.name] = None
        else:
            bcut = None
            bval = INF
            for c in cuts[n.name]:
                if c == frozenset([n.name]):
                    continue
                s = 0
                for m in c:
                    s += cost.get(m,0)
                cval = s + 1
                if cval < bval:
                    bval = cval; bcut = c
            if bcut is None:
                fl = frozenset([f.name for f in n.fanins])
                bcut = fl
                s = 0
                for m in bcut:
                    s += cost.get(m,0)
                bval = s + 1
            cost[n.name] = bval
            best_cut[n.name] = bcut
    mapping_nodes = set()
    dq = deque()
    for out in output_list:
        if out in best_cut and best_cut[out] is not None:
            dq.append(out)
    while dq:
        u = dq.pop()
        if u in mapping_nodes:
            continue
        mapping_nodes.add(u)
        c = best_cut[u]
        if c:
            for m in c:
                if m in nodes:
                    nm = nodes[m]
                    if m not in input_list and nm.const_val is None:
                        dq.append(m)
    mapping_topo = [n for n in topo_list if n.name in mapping_nodes]
    f = open(solution_file,'w')
    f.write('.model '+model_name+'\n')
    f.write('.inputs '+' '.join(input_list)+'\n')
    f.write('.outputs '+' '.join(output_list)+'\n')
    for n in mapping_topo:
        leaves = best_cut[n.name]
        inputs = sorted(leaves)
        k = len(inputs)
        N = 1<<k
        full_mask = (1<<N) - 1
        masks = []
        for i in range(k):
            m = 0
            for j in range(N):
                if (j>>i)&1:
                    m |= (1<<j)
            masks.append(m)
        vis = set()
        queue2 = [n.name]
        while queue2:
            u = queue2.pop()
            if u in vis or u in inputs:
                continue
            vis.add(u)
            for fn in nodes[u].fanins:
                if fn.name not in vis:
                    queue2.append(fn.name)
        cone_nodes = [nodes[name] for name in topo_names if name in vis]
        val = {}
        for i,name in enumerate(inputs):
            val[name] = masks[i]
        for v in cone_nodes:
            if v.const_val is not None:
                val[v.name] = full_mask if v.const_val==1 else 0
            else:
                if v.patterns_zero:
                    zero_mask = 0
                    for p in v.patterns_zero:
                        m = full_mask
                        for idx,ch in enumerate(p):
                            vm = val[v.fanin_names[idx]]
                            if ch=='1':
                                m &= vm
                            elif ch=='0':
                                m &= (~vm)&full_mask
                        zero_mask |= m
                    val[v.name] = (~zero_mask)&full_mask
                else:
                    one_mask = 0
                    for p in v.patterns_one:
                        m = full_mask
                        for idx,ch in enumerate(p):
                            vm = val[v.fanin_names[idx]]
                            if ch=='1':
                                m &= vm
                            elif ch=='0':
                                m &= (~vm)&full_mask
                        one_mask |= m
                    val[v.name] = one_mask
        root_mask = val[n.name]
        f.write('.names '+' '.join(inputs)+' '+n.name+'\n')
        for j in range(N):
            if (root_mask>>j)&1:
                line = ''.join('1' if (j>>i)&1 else '0' for i in range(k))
                f.write(line+' 1\n')
    f.write('.end\n')
    f.close()

\end{minted}

In the second iteration, the model overcorrects by adopting a naive 6-LUT mapping strategy that trivially assigns each node to a separate LUT. While this approach passes all test cases within the timeout, it produces poor solution quality. Specifically, it redundantly maps each non-primary-input node as an individual LUT by directly copying inputs to outputs, resulting in excessive and inefficient resource usage.

\begin{minted}[breaklines=true, breakanywhere=true,fontsize=\scriptsize, linenos, frame=single, highlightcolor=yellow!35, highlightlines={28-30}]{python}
# iteration 2: a very naive mapping strategy. Simply copy the input to the output. 

def solve(input_file: str, solution_file: str):
    raw = open(input_file).read().splitlines()
    merged = []
    i = 0
    while i < len(raw):
        line = raw[i].rstrip()
        if not line or line.lstrip().startswith('#'):
            i += 1
            continue
        if line.endswith('\\'):
            accum = line[:-1].rstrip()
            i += 1
            while i < len(raw):
                nxt = raw[i].rstrip()
                if nxt.endswith('\\'):
                    accum += ' ' + nxt[:-1].rstrip()
                    i += 1
                else:
                    accum += ' ' + nxt
                    i += 1
                    break
            merged.append(accum)
        else:
            merged.append(line)
            i += 1
    with open(solution_file, 'w') as f:
        for l in merged:
            f.write(l + '\n')
\end{minted}

In the third iteration, the model proposes a more sophisticated strategy that first applies static pruning to the cut enumeration space -- limiting the number of cuts per node to 20 -- followed by a dynamic programming-based cut selection. This approach improves solution quality, though full yield is still not achieved. Additionally, truth table computation is optimized using bitwise operations.

\begin{minted}[breaklines=true, breakanywhere=true,fontsize=\scriptsize, linenos, frame=single, highlightcolor=yellow!35, highlightlines={138,151-156,185-189}]{python}
# iteration 3: static pruning by capping the number of cuts per node to M = 20; efficient bitwise operations

def solve(input_file: str, solution_file: str):
    from collections import deque
    raw_lines = open(input_file).read().splitlines()
    lines = []
    i = 0
    while i < len(raw_lines):
        line = raw_lines[i].rstrip()
        if not line or line.lstrip().startswith('#'):
            i += 1
            continue
        if line.endswith('\\'):
            accum = line[:-1].rstrip()
            i += 1
            while i < len(raw_lines):
                nxt = raw_lines[i].rstrip()
                if nxt.endswith('\\'):
                    accum += ' ' + nxt[:-1].rstrip()
                    i += 1
                else:
                    accum += ' ' + nxt
                    i += 1
                    break
            lines.append(accum)
        else:
            lines.append(line)
            i += 1
    class Node:
        def __init__(self, name):
            self.name = name
            self.fanin_names = []
            self.patterns_one = []
            self.patterns_zero = []
            self.const_val = None
            self.fanins = []
            self.fanouts = []
    model_name = ''
    input_list = []
    output_list = []
    nodes = {}
    idx = 0
    while idx < len(lines):
        l = lines[idx].strip()
        if not l:
            idx += 1
            continue
        if l.startswith('.model'):
            parts = l.split()
            if len(parts) > 1:
                model_name = parts[1]
            idx += 1
        elif l.startswith('.inputs'):
            parts = l.split()
            for name in parts[1:]:
                input_list.append(name)
                if name not in nodes:
                    nodes[name] = Node(name)
            idx += 1
        elif l.startswith('.outputs'):
            parts = l.split()
            for name in parts[1:]:
                output_list.append(name)
            idx += 1
        elif l.startswith('.names'):
            parts = l.split()
            in_names = parts[1:-1]
            out_name = parts[-1]
            if out_name not in nodes:
                nodes[out_name] = Node(out_name)
            node = nodes[out_name]
            node.fanin_names = in_names[:]
            patterns = []
            j = idx + 1
            while j < len(lines) and not lines[j].startswith('.'):
                s = lines[j].strip()
                if s:
                    patterns.append(s)
                j += 1
            if len(in_names) == 0:
                if patterns and patterns[0] == '1':
                    node.const_val = 1
                else:
                    node.const_val = 0
            else:
                for pline in patterns:
                    parts2 = pline.split()
                    if len(parts2) == 1:
                        mask = parts2[0]; bit = '1'
                    else:
                        mask = parts2[0]; bit = parts2[1]
                    if bit == '1':
                        node.patterns_one.append(mask)
                    else:
                        node.patterns_zero.append(mask)
            idx = j
        elif l.startswith('.end'):
            break
        else:
            idx += 1
    for node in nodes.values():
        node.fanins = [nodes[nm] for nm in node.fanin_names if nm in nodes]
        node.fanouts = []
    for node in nodes.values():
        for f in node.fanins:
            f.fanouts.append(node)
    reachable = set()
    stack = [nm for nm in output_list if nm in nodes]
    while stack:
        u = stack.pop()
        if u in reachable: continue
        reachable.add(u)
        for p in nodes[u].fanin_names:
            if p in nodes and p not in reachable:
                stack.append(p)
    nodes = {nm:nd for nm,nd in nodes.items() if nm in reachable}
    for nd in nodes.values():
        nd.fanins = [f for f in nd.fanins if f.name in nodes]
        nd.fanouts = [f for f in nd.fanouts if f.name in nodes]
    indeg = {nm: len(nd.fanins) for nm,nd in nodes.items()}
    dq = deque([nm for nm,d in indeg.items() if d == 0])
    topo = []
    while dq:
        u = dq.popleft()
        topo.append(u)
        for w in nodes[u].fanouts:
            indeg[w.name] -= 1
            if indeg[w.name] == 0:
                dq.append(w.name)
    topo_nodes = [nodes[nm] for nm in topo]
    name_to_idx = {nd.name:i for i,nd in enumerate(topo_nodes)}
    idx_to_node = topo_nodes
    idx_to_name = [nd.name for nd in topo_nodes]
    N = len(topo_nodes)
    PI_idx = set(name_to_idx[nm] for nm in input_list if nm in name_to_idx)
    self_mask = [1 << i for i in range(N)]
    K = 6
    M = 20
    cuts = [[] for _ in range(N)]
    for i, nd in enumerate(topo_nodes):
        if i in PI_idx or nd.const_val is not None:
            cuts[i] = [self_mask[i]]
            continue
        fan_idxs = [name_to_idx[x] for x in nd.fanin_names if x in name_to_idx]
        c_list = None
        for f in fan_idxs:
            fcuts = cuts[f]
            if c_list is None:
                c_list = fcuts[:M]
            else:
                newset = set()
                for a in c_list[:M]:
                    for b in fcuts[:M]:
                        u = a | b
                        if u.bit_count() <= K:
                            newset.add(u)
                if newset:
                    lst = sorted(newset, key=lambda x: x.bit_count())
                    c_list = lst[:M]
                else:
                    c_list = []
            if not c_list:
                break
        if c_list is None:
            c_list = []
        s = set(c_list)
        s.add(self_mask[i])
        lst2 = sorted(s, key=lambda x: x.bit_count())
        cuts[i] = lst2[:M]
    INF = 10**18
    cost = [0]*N
    bestM = [None]*N
    for i, nd in enumerate(topo_nodes):
        if i in PI_idx or nd.const_val is not None:
            cost[i] = 0
            bestM[i] = None
            continue
        bestc = INF
        bm = None
        for cm in cuts[i]:
            if cm == self_mask[i]:
                continue
            s = 0
            m = cm
            while m:
                lsb = m & -m
                j = lsb.bit_length() - 1
                s += cost[j]
                m ^= lsb
            cval = s + 1
            if cval < bestc:
                bestc = cval
                bm = cm
        if bm is None:
            s = 0
            mask = 0
            for f in nd.fanin_names:
                if f in name_to_idx:
                    j = name_to_idx[f]
                    mask |= self_mask[j]
                    s += cost[j]
            bm = mask
            bestc = s + 1
        cost[i] = bestc
        bestM[i] = bm
    mapping = set()
    stack = [name_to_idx[nm] for nm in output_list if nm in name_to_idx]
    visited = set()
    while stack:
        u = stack.pop()
        if u in visited:
            continue
        visited.add(u)
        bm = bestM[u]
        if bm is None or bm == self_mask[u]:
            continue
        mapping.add(u)
        m = bm
        while m:
            lsb = m & -m
            j = lsb.bit_length() - 1
            m ^= lsb
            if j not in visited and j not in PI_idx and topo_nodes[j].const_val is None:
                stack.append(j)
    mapped = [i for i in range(N) if i in mapping]
    with open(solution_file, 'w') as f:
        f.write('.model ' + model_name + '\n')
        f.write('.inputs ' + ' '.join(input_list) + '\n')
        f.write('.outputs ' + ' '.join(output_list) + '\n')
        for i in mapped:
            nd = topo_nodes[i]
            bm = bestM[i]
            leaves = [j for j in range(N) if (bm >> j) & 1]
            inputs = [idx_to_name[j] for j in leaves]
            k = len(inputs)
            Np = 1 << k
            full = (1 << Np) - 1
            masks = []
            for t in range(k):
                mm = 0
                for j in range(Np):
                    if (j >> t) & 1:
                        mm |= 1 << j
                masks.append(mm)
            val = {}
            for t,j in enumerate(leaves):
                val[j] = masks[t]
            vis = set(leaves)
            stack2 = [i]
            while stack2:
                u = stack2.pop()
                if u in vis:
                    continue
                vis.add(u)
                for p in topo_nodes[u].fanins:
                    pi = name_to_idx.get(p.name)
                    if pi is not None and pi not in vis:
                        stack2.append(pi)
            cone = [j for j in range(N) if j in vis and j not in leaves]
            for j in sorted(cone, key=lambda x: name_to_idx[topo_nodes[x].name]):
                v = topo_nodes[j]
                if v.const_val is not None:
                    val[j] = full if v.const_val == 1 else 0
                else:
                    if v.patterns_zero:
                        zm = 0
                        for p in v.patterns_zero:
                            mm = full
                            for t,ch in enumerate(p):
                                fn = v.fanin_names[t]
                                pj = name_to_idx.get(fn)
                                if pj is None:
                                    mm = 0
                                    break
                                vm = val[pj]
                                if ch == '1':
                                    mm &= vm
                                elif ch == '0':
                                    mm &= (~vm) & full
                            zm |= mm
                        val[j] = (~zm) & full
                    else:
                        om = 0
                        for p in v.patterns_one:
                            mm = full
                            for t,ch in enumerate(p):
                                fn = v.fanin_names[t]
                                pj = name_to_idx.get(fn)
                                if pj is None:
                                    mm = 0
                                    break
                                vm = val[pj]
                                if ch == '1':
                                    mm &= vm
                                elif ch == '0':
                                    mm &= (~vm) & full
                            om |= mm
                        val[j] = om
            root = val[i]
            f.write('.names ' + ' '.join(inputs) + ' ' + nd.name + '\n')
            for j in range(Np):
                if (root >> j) & 1:
                    bits = ''.join('1' if (j >> t) & 1 else '0' for t in range(k))
                    f.write(bits + ' 1\n')
        f.write('.end\n')

\end{minted}

In the fourth iteration, the cut limit per node is increased from 20 to 30, enabling broader solution exploration and potentially improving quality. Additionally, the algorithm reduces redundant computations by caching precomputed scores for each cut.

\begin{minted}[breaklines=true, breakanywhere=true,fontsize=\scriptsize, linenos, frame=single,  highlightcolor=yellow!35, highlightlines={126, 143-155, 168-173}]{python}
# iteration 4: explore larger solution space; reduce redundant computations

def solve(input_file: str, solution_file: str):
    from collections import deque
    raw = open(input_file).read().splitlines()
    lines = []
    i = 0
    while i < len(raw):
        l = raw[i].rstrip()
        if not l or l.lstrip().startswith('#'):
            i += 1
            continue
        if l.endswith('\\'):
            acc = l[:-1].rstrip()
            i += 1
            while i < len(raw):
                nl = raw[i].rstrip()
                if nl.endswith('\\'):
                    acc += ' ' + nl[:-1].rstrip()
                    i += 1
                else:
                    acc += ' ' + nl
                    i += 1
                    break
            lines.append(acc)
        else:
            lines.append(l)
            i += 1
    class Node:
        __slots__ = ('name','fanin_names','patterns_one','patterns_zero','const_val','fanins','fanouts')
        def __init__(self,n):
            self.name = n
            self.fanin_names = []
            self.patterns_one = []
            self.patterns_zero = []
            self.const_val = None
            self.fanins = []
            self.fanouts = []
    model = ''
    inputs = []
    outputs = []
    nodes = {}
    idx = 0
    while idx < len(lines):
        l = lines[idx].strip()
        if not l:
            idx += 1; continue
        if l.startswith('.model'):
            parts = l.split()
            if len(parts)>1: model = parts[1]
            idx += 1
        elif l.startswith('.inputs'):
            parts = l.split()
            for nm in parts[1:]:
                inputs.append(nm)
                if nm not in nodes: nodes[nm] = Node(nm)
            idx += 1
        elif l.startswith('.outputs'):
            parts = l.split()
            for nm in parts[1:]:
                outputs.append(nm)
            idx += 1
        elif l.startswith('.names'):
            parts = l.split()
            inps = parts[1:-1]; outp = parts[-1]
            if outp not in nodes: nodes[outp] = Node(outp)
            nd = nodes[outp]
            nd.fanin_names = inps[:]
            pats = []
            j = idx+1
            while j < len(lines) and not lines[j].startswith('.'):
                s = lines[j].strip()
                if s: pats.append(s)
                j += 1
            if not inps:
                if pats and pats[0]=='1': nd.const_val = 1
                else: nd.const_val = 0
            else:
                for pt in pats:
                    sp = pt.split()
                    if len(sp)==1:
                        mask = sp[0]; bit = '1'
                    else:
                        mask,bit = sp[0],sp[1]
                    if bit=='1': nd.patterns_one.append(mask)
                    else: nd.patterns_zero.append(mask)
            idx = j
        elif l.startswith('.end'):
            break
        else:
            idx += 1
    for nd in nodes.values():
        nd.fanins = [nodes[nm] for nm in nd.fanin_names if nm in nodes]
    for nd in nodes.values():
        for f in nd.fanins:
            f.fanouts.append(nd)
    reachable = set()
    st = [nm for nm in outputs if nm in nodes]
    while st:
        u = st.pop()
        if u in reachable: continue
        reachable.add(u)
        for p in nodes[u].fanin_names:
            if p in nodes and p not in reachable:
                st.append(p)
    nodes = {nm:nd for nm,nd in nodes.items() if nm in reachable}
    for nd in nodes.values():
        nd.fanins = [f for f in nd.fanins if f.name in nodes]
        nd.fanouts = [f for f in nd.fanouts if f.name in nodes]
    indeg = {nm: len(nd.fanins) for nm,nd in nodes.items()}
    dq = deque([nm for nm,d in indeg.items() if d==0])
    topo = []
    while dq:
        u = dq.popleft(); topo.append(u)
        for w in nodes[u].fanouts:
            indeg[w.name] -= 1
            if indeg[w.name]==0: dq.append(w.name)
    topo_nodes = [nodes[nm] for nm in topo]
    N = len(topo_nodes)
    name_to_idx = {nd.name:i for i,nd in enumerate(topo_nodes)}
    idx_to_name = [nd.name for nd in topo_nodes]
    PI = set(name_to_idx[nm] for nm in inputs if nm in name_to_idx)
    cost = [0]*N
    bestM = [None]*N
    K = 6
    M = 30
    cuts = [[] for _ in range(N)]
    for i,nd in enumerate(topo_nodes):
        if i in PI or nd.const_val is not None:
            cost[i] = 0
            bestM[i] = None
            cuts[i] = [(1<<i, 0)]
            continue
        fans = [name_to_idx[nm] for nm in nd.fanin_names if nm in name_to_idx]
        c_list = None
        for f in fans:
            fcuts = cuts[f]
            if len(fcuts) > M: fcuts = fcuts[:M]
            if c_list is None:
                c_list = fcuts.copy()
            else:
                newm = {}
                for m1,sc1 in c_list:
                    for m2,sc2 in fcuts:
                        m = m1 | m2
                        if m.bit_count() <= K:
                            if m in newm: continue
                            t = m; sc = 0
                            while t:
                                lsb = t & -t; j = lsb.bit_length()-1
                                sc += cost[j]; t ^= lsb
                            newm[m] = sc
                if not newm:
                    c_list = []
                    break
                items = sorted(newm.items(), key=lambda x: x[1])
                c_list = items[:M]
        if c_list is None: c_list = []
        fb = 0
        scf = 0
        for f in fans:
            fb |= (1<<f)
            scf += cost[f]
        if fb.bit_count() <= K:
            if not any(m==fb for m,_ in c_list):
                c_list.append((fb, scf))
        bestc = 10**18; bm = None
        for m,sc in c_list:
            v = sc + 1
            if v < bestc:
                bestc = v; bm = m
        if bm is None:
            bm = fb; bestc = scf + 1
        cost[i] = bestc; bestM[i] = bm
        cuts[i] = sorted(c_list, key=lambda x: x[1])[:M]
    mapping = set()
    st = [name_to_idx[nm] for nm in outputs if nm in name_to_idx]
    vis = set()
    while st:
        u = st.pop()
        if u in vis: continue
        vis.add(u)
        bm = bestM[u]
        if bm is None: continue
        mapping.add(u)
        t = bm
        while t:
            lsb = t & -t; j = lsb.bit_length()-1; t ^= lsb
            if j in PI or topo_nodes[j].const_val is not None: continue
            if j not in vis: st.append(j)
    mapped = sorted(mapping)
    with open(solution_file, 'w') as f:
        f.write('.model ' + model + '\n')
        f.write('.inputs ' + ' '.join(inputs) + '\n')
        f.write('.outputs ' + ' '.join(outputs) + '\n')
        for i in mapped:
            nd = topo_nodes[i]
            bm = bestM[i]
            leaves = [j for j in range(N) if (bm>>j)&1]
            inputs_l = [idx_to_name[j] for j in leaves]
            k = len(inputs_l)
            Np = 1<<k
            full = (1<<Np) - 1
            masks = []
            for t in range(k):
                mm = 0
                for j in range(Np):
                    if (j>>t)&1: mm |= 1<<j
                masks.append(mm)
            val = {}
            for t,j in enumerate(leaves):
                val[j] = masks[t]
            vis2 = set(leaves)
            stk = [i]
            while stk:
                u = stk.pop()
                if u in vis2: continue
                vis2.add(u)
                for p in topo_nodes[u].fanins:
                    pj = name_to_idx.get(p.name)
                    if pj is not None and pj not in vis2:
                        stk.append(pj)
            cone = [j for j in topo if False]
            cone = [j for j in range(N) if j in vis2 and j not in leaves]
            for j in sorted(cone):
                v = topo_nodes[j]
                if v.const_val is not None:
                    val[j] = full if v.const_val==1 else 0
                else:
                    if v.patterns_zero:
                        zm = 0
                        for p in v.patterns_zero:
                            mm = full
                            for t,ch in enumerate(p):
                                fn = v.fanin_names[t]
                                pj = name_to_idx.get(fn)
                                if pj is None:
                                    mm = 0; break
                                vm = val[pj]
                                if ch=='1': mm &= vm
                                elif ch=='0': mm &= (~vm)&full
                            zm |= mm
                        val[j] = (~zm)&full
                    else:
                        om = 0
                        for p in v.patterns_one:
                            mm = full
                            for t,ch in enumerate(p):
                                fn = v.fanin_names[t]
                                pj = name_to_idx.get(fn)
                                if pj is None:
                                    mm = 0; break
                                vm = val[pj]
                                if ch=='1': mm &= vm
                                elif ch=='0': mm &= (~vm)&full
                            om |= mm
                        val[j] = om
            root = val[i]
            f.write('.names ' + ' '.join(inputs_l) + ' ' + nd.name + '\n')
            for j in range(Np):
                if (root>>j)&1:
                    pat = ''.join('1' if (j>>t)&1 else '0' for t in range(k))
                    f.write(pat + ' 1\n')
        f.write('.end\n')

\end{minted}

In the fifth iteration, it introduces a dynamic pruning method by adapting the cut limit based on circuit size---more cuts for smaller circuits (higher quality), and fewer for large circuits (better yield).

\begin{minted}[breaklines=true, breakanywhere=true,fontsize=\scriptsize, linenos, frame=single, highlightlines={125-132}, highlightcolor=yellow!35]{python}
# iteration 5: dynamic pruning of cut enumeration space

def solve(input_file: str, solution_file: str):
    from collections import deque
    raw = open(input_file).read().splitlines()
    lines = []
    i = 0
    while i < len(raw):
        l = raw[i].rstrip()
        if not l or l.lstrip().startswith('#'):
            i += 1
            continue
        if l.endswith('\\'):
            acc = l[:-1].rstrip()
            i += 1
            while i < len(raw):
                nl = raw[i].rstrip()
                if nl.endswith('\\'):
                    acc += ' ' + nl[:-1].rstrip()
                    i += 1
                else:
                    acc += ' ' + nl
                    i += 1
                    break
            lines.append(acc)
        else:
            lines.append(l)
            i += 1
    class Node:
        __slots__ = ('name','fanin_names','patterns_one','patterns_zero','const_val','fanins','fanouts')
        def __init__(self,n):
            self.name = n
            self.fanin_names = []
            self.patterns_one = []
            self.patterns_zero = []
            self.const_val = None
            self.fanins = []
            self.fanouts = []
    model = ''
    inputs = []
    outputs = []
    nodes = {}
    idx = 0
    while idx < len(lines):
        l = lines[idx].strip()
        if not l:
            idx += 1; continue
        if l.startswith('.model'):
            parts = l.split()
            if len(parts) > 1: model = parts[1]
            idx += 1
        elif l.startswith('.inputs'):
            parts = l.split()
            for nm in parts[1:]:
                inputs.append(nm)
                if nm not in nodes: nodes[nm] = Node(nm)
            idx += 1
        elif l.startswith('.outputs'):
            parts = l.split()
            for nm in parts[1:]:
                outputs.append(nm)
            idx += 1
        elif l.startswith('.names'):
            parts = l.split()
            inps = parts[1:-1]; outp = parts[-1]
            if outp not in nodes: nodes[outp] = Node(outp)
            nd = nodes[outp]
            nd.fanin_names = inps[:]
            pats = []
            j = idx + 1
            while j < len(lines) and not lines[j].startswith('.'):
                s = lines[j].strip()
                if s: pats.append(s)
                j += 1
            if not inps:
                nd.const_val = 1 if pats and pats[0] == '1' else 0
            else:
                for pt in pats:
                    sp = pt.split()
                    if len(sp) == 1:
                        mask, bit = sp[0], '1'
                    else:
                        mask, bit = sp[0], sp[1]
                    if bit == '1': nd.patterns_one.append(mask)
                    else: nd.patterns_zero.append(mask)
            idx = j
        elif l.startswith('.end'):
            break
        else:
            idx += 1
    for nd in nodes.values():
        nd.fanins = [nodes[nm] for nm in nd.fanin_names if nm in nodes]
    for nd in nodes.values():
        for f in nd.fanins:
            f.fanouts.append(nd)
    reachable = set()
    st = [nm for nm in outputs if nm in nodes]
    while st:
        u = st.pop()
        if u in reachable: continue
        reachable.add(u)
        for p in nodes[u].fanin_names:
            if p in nodes and p not in reachable:
                st.append(p)
    nodes = {nm:nd for nm,nd in nodes.items() if nm in reachable}
    for nd in nodes.values():
        nd.fanins = [f for f in nd.fanins if f.name in nodes]
        nd.fanouts = [f for f in nd.fanouts if f.name in nodes]
    indeg = {nm: len(nd.fanins) for nm,nd in nodes.items()}
    dq = deque([nm for nm,d in indeg.items() if d == 0])
    topo = []
    while dq:
        u = dq.popleft(); topo.append(u)
        for w in nodes[u].fanouts:
            indeg[w.name] -= 1
            if indeg[w.name] == 0: dq.append(w.name)
    topo_nodes = [nodes[nm] for nm in topo]
    N = len(topo_nodes)
    name_to_idx = {nd.name: i for i, nd in enumerate(topo_nodes)}
    idx_to_name = [nd.name for nd in topo_nodes]
    PI = set(name_to_idx[nm] for nm in inputs if nm in name_to_idx)
    cost = [0] * N
    bestM = [None] * N
    K = 6
    if N <= 1500:
        M = 64
    elif N <= 3000:
        M = 48
    elif N <= 5000:
        M = 32
    else:
        M = 20
    cuts = [[] for _ in range(N)]
    def prune_cuts(items, limit):
        items_sorted = sorted(items, key=lambda x: (x[1], x[0].bit_count()))
        pr = []
        for m, sc in items_sorted:
            dom = False
            for pm, psc in pr:
                if psc <= sc and (pm & m) == pm:
                    dom = True
                    break
            if not dom:
                pr.append((m, sc))
                if len(pr) >= limit:
                    break
        return pr
    for i, nd in enumerate(topo_nodes):
        if i in PI or nd.const_val is not None:
            cost[i] = 0
            bestM[i] = None
            cuts[i] = [(1 << i, 0)]
            continue
        fans = [name_to_idx[nm] for nm in nd.fanin_names if nm in name_to_idx]
        fans.sort(key=lambda x: len(cuts[x]))
        c_list = None
        for f in fans:
            fcuts = cuts[f]
            if not fcuts:
                c_list = []
                break
            fcuts = fcuts[:M]
            if c_list is None:
                c_list = fcuts.copy()
            else:
                newm = {}
                for m1, sc1 in c_list:
                    for m2, sc2 in fcuts:
                        m = m1 | m2
                        if m.bit_count() <= K:
                            s2 = sc1 + sc2
                            prev = newm.get(m)
                            if prev is None or s2 < prev:
                                newm[m] = s2
                if not newm:
                    c_list = []
                    break
                c_list = prune_cuts(list(newm.items()), M)
        if not c_list:
            um = 0; usc = 0
            for f in fans:
                um |= (1 << f)
                usc += cost[f]
            c_list = [(um, usc)]
        um = 0; usc = 0
        for f in fans:
            um |= (1 << f)
            usc += cost[f]
        if um.bit_count() <= K and all(m != um for m, _ in c_list):
            c_list.append((um, usc))
        c_list = prune_cuts(c_list, M)
        bestc = 10**18; bm = None
        selfm = (1 << i)
        for m, sc in c_list:
            if m == selfm:
                continue
            v = sc + 1
            if v < bestc:
                bestc = v; bm = m
        if bm is None:
            bm = um
            bestc = usc + 1
        cost[i] = bestc
        bestM[i] = bm
        cuts[i] = c_list
    mapping = set()
    st = [name_to_idx[nm] for nm in outputs if nm in name_to_idx]
    vis = set()
    while st:
        u = st.pop()
        if u in vis: continue
        vis.add(u)
        bm = bestM[u]
        if bm is None: continue
        mapping.add(u)
        t = bm
        while t:
            lsb = t & -t
            j = lsb.bit_length() - 1
            t ^= lsb
            if j in PI or topo_nodes[j].const_val is not None: continue
            if j not in vis:
                st.append(j)
    mapped = sorted(mapping)
    with open(solution_file, 'w') as f:
        f.write('.model ' + model + '\n')
        f.write('.inputs ' + ' '.join(inputs) + '\n')
        f.write('.outputs ' + ' '.join(outputs) + '\n')
        for i in mapped:
            nd = topo_nodes[i]
            bm = bestM[i]
            leaves = [j for j in range(N) if (bm >> j) & 1]
            inputs_l = [idx_to_name[j] for j in leaves]
            k = len(inputs_l)
            Np = 1 << k
            full = (1 << Np) - 1
            masks = []
            for t in range(k):
                mm = 0
                for j in range(Np):
                    if (j >> t) & 1:
                        mm |= 1 << j
                masks.append(mm)
            val = {}
            for t, j in enumerate(leaves):
                val[j] = masks[t]
            seen = set(leaves)
            stk = [i]
            cone = []
            while stk:
                u = stk.pop()
                if u in seen: continue
                seen.add(u)
                cone.append(u)
                for p in topo_nodes[u].fanins:
                    pj = name_to_idx.get(p.name)
                    if pj is not None and pj not in seen:
                        stk.append(pj)
            cone.sort()
            for j in cone:
                v = topo_nodes[j]
                if v.const_val is not None:
                    val[j] = full if v.const_val == 1 else 0
                else:
                    if v.patterns_zero:
                        zm = 0
                        for p in v.patterns_zero:
                            mm = full
                            for t, ch in enumerate(p):
                                fn = v.fanin_names[t]
                                pj = name_to_idx.get(fn)
                                if pj is None:
                                    mm = 0
                                    break
                                vm = val[pj]
                                if ch == '1':
                                    mm &= vm
                                elif ch == '0':
                                    mm &= (~vm) & full
                            zm |= mm
                        val[j] = (~zm) & full
                    else:
                        om = 0
                        for p in v.patterns_one:
                            mm = full
                            for t, ch in enumerate(p):
                                fn = v.fanin_names[t]
                                pj = name_to_idx.get(fn)
                                if pj is None:
                                    mm = 0
                                    break
                                vm = val[pj]
                                if ch == '1':
                                    mm &= vm
                                elif ch == '0':
                                    mm &= (~vm) & full
                            om |= mm
                        val[j] = om
            root = val[i]
            f.write('.names ' + ' '.join(inputs_l) + ' ' + nd.name + '\n')
            for j in range(Np):
                if (root >> j) & 1:
                    pat = ''.join('1' if (j >> t) & 1 else '0' for t in range(k))
                    f.write(pat + ' 1\n')
        f.write('.end\n')

\end{minted}

\section{Datasets}
\label{appendix:dataset}

We summarize the original data sources for each problem in Table~\ref{tab:dataset} and the number of data instances in Table~\ref{tab:instances}.
All datasets are derived from real-world applications.
We further partition or transform them into standardized input formats, ensuring the inclusion of both small-scale instances for demonstration purposes and large-scale instances for evaluation.
For detailed data organization, please refer to our repository.

\begin{table}[!htbp]
\caption{Datasets used in our benchmark.}
\label{tab:dataset}
\centering
\resizebox{\linewidth}{!}{
\begin{tabular}{ll}\hline
\textbf{Problem} & \textbf{Original Data Source}\\\hline
Operator scheduling & EXPRESS~\citep{wang2007express}\\
Technology mapping & EPFL~\citep{amaru2015epfl} and ISCAS85~\citep{hansen1999iscas}\\
Global routing & ISPD'24 Contest~\citep{liang2024ispdcontest}\\\hline
E-graph extraction & SmoothE~\citep{cai2025egraph}\\
Intra-op parallelism & ASPLOS'24 Contest~\citep{moffit2025iopddl}\\\hline
Protein sequence design & Protein Data Bank (PDB)~\citep{pdb}\\
Mendelian error detection & Cost Function Library~\citep{sanchez2008mendelian,schiex_cost_function_library}\\\hline
Airline crew pairing & China Graduate Mathematical Modeling Competition'21 F~\citep{cmmc}\\
Pickup and delivery w/ time windows & MetaPDPTW~\citep{li2001metapdptw}\\\hline
\end{tabular}
}
\end{table}
\begin{table}[!htbp]
\caption{Number of instances of each problem in \Name.}
\label{tab:instances}
\centering
\small
\begin{tabular}{ll}\hline
\textbf{Problem} & \textbf{\# of Instances}\\\hline
Operator scheduling & 24\\
Technology mapping & 31\\
Global routing & 24\\
E-graph extraction & 23\\
Intra-op parallelism & 28\\
Protein sequence design & 24\\
Mendelian error detection & 20\\
Airline crew pairing & 14\\
Pickup and delivery w/ time windows & 30\\
\textbf{Total} & 218\\\hline
\end{tabular}
\end{table}

\end{document}